\documentclass[conference,dvipsnames]{IEEEtran}

\usepackage[T1]{fontenc}
\usepackage{times}
\usepackage{amsmath, amssymb, amsfonts}
\usepackage{dsfont}           
\usepackage{booktabs}         
\usepackage{multirow}         
\usepackage{graphicx}         
\usepackage{xcolor}           
\usepackage{soul}             
\usepackage{wrapfig}          
\usepackage{float}            
\usepackage{stfloats}         

\usepackage{algorithm}        
\usepackage{algorithmic}      

\usepackage{subcaption}       
\usepackage{caption}          
\usepackage{tabularx}         
\usepackage{hhline}           

\usepackage{listings}         

\usepackage[square, numbers, sort&compress]{natbib}

\usepackage[bookmarks=true]{hyperref}
  \hypersetup{
    colorlinks   = true,
    linkcolor    = orange,
    citecolor    = orange,
    filecolor    = orange,
    urlcolor     = orange
  }
\usepackage[capitalise,nameinlink]{cleveref}  

\usepackage{comment}
\usepackage{footnote}         
\usepackage{xspace}           

\newcommand{\MethodName}{Mid‐Level MoE}

\newcommand{\insertteaser}{
  \centering
  \includegraphics[width=0.99\linewidth]{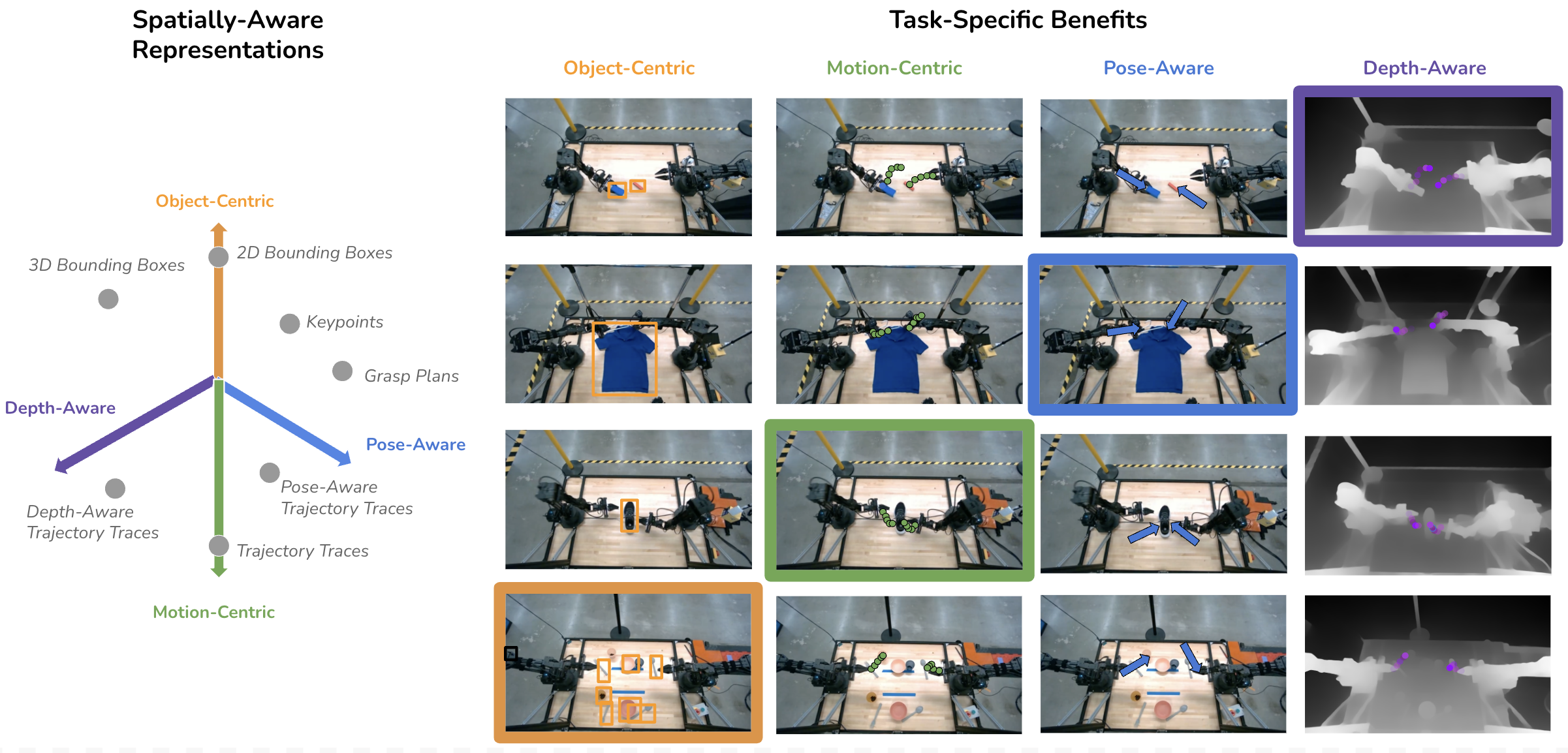} \\
  \captionof{figure}{\textbf{Bimanual, dexterous manipulation requires task‐specific grounding.} 
    The left depicts various axes for spatial grounding as well as qualitative categorizations of different mid‐level representations. 
    Different representations lead to different levels of improvement depending on the task.}
  \label{fig:whatever}
}

\makeatletter
\apptocmd{\@maketitle}{\centering\insertteaser}{}{}
\makeatother

\begin{document}

\title{Bridging Perception and Action: Spatially-Grounded Mid-Level Representations for Robot Generalization}

\author{Jonathan Yang\IEEEauthorrefmark{1}\IEEEauthorrefmark{2}, Chuyuan Kelly Fu \IEEEauthorrefmark{2},  Dhruv Shah\IEEEauthorrefmark{2}, Dorsa Sadigh\IEEEauthorrefmark{1}\IEEEauthorrefmark{2}, \\ Fei Xia\IEEEauthorrefmark{2}, Tingnan Zhang\IEEEauthorrefmark{2} \\
\IEEEauthorblockA{\IEEEauthorrefmark{1} Stanford University}
\IEEEauthorblockA{\IEEEauthorrefmark{2} Google DeepMind}}


%

\maketitle

\begin{abstract}
%
%
%

In this work, we investigate how \textit{spatially-grounded} auxiliary representations can provide both broad, high-level grounding, as well as direct, actionable information to help policy learning performance and generalization for dexterous tasks. We study these mid-level representations across three critical dimensions: object-centricity, pose-awareness, and depth-awareness. 
We use these interpretable mid-level representations to train \emph{specialist} encoders via supervised learning, then use these representations as inputs to a diffusion policy to solve dexterous bimanual manipulation tasks in the real-world.
%
We propose a novel mixture-of-experts policy architecture that can combine multiple specialized expert models, each trained on a distinct mid-level representation, to improve the generalization of the policy. This method achieves an average of $11\%$ higher success rate on average over a language-grounded baseline and a $24\%$ higher success rate over a standard diffusion policy baseline for our evaluation tasks. Furthermore, we find that leveraging mid-level representations as supervision signals for policy actions within a weighted imitation learning algorithm improves the precision with which the policy follows these representations, leading to an additional performance increase of $10\%$.
%
%
Our findings highlight the importance of grounding robot policies with not only broad, perceptual tasks, but also more granular, actionable representations. For further information and videos, please visit \url{https://mid-level-moe.github.io}.
\end{abstract}

\IEEEpeerreviewmaketitle

\section{Introduction}

Large pre-trained robotics models have made significant progress in recent years towards improving robotic generalization capabilities by leveraging large-scale pre-training datasets. However, these models still face challenges in adapting to slight scene variations such as different spatial locations, unseen objects, and different lighting conditions. An increasingly popular approach to address this challenge is explicitly establishing deeper connections between robot policies and the abstract patterns and relationships that govern the physical world.  For example, vision-language-action models (VLAs) make an attempt to benefit from semantic and visual knowledge from vision-language models (VLMs) by fine-tuning these models with robot data. Other works use explicit \emph{mid-level representations} such as low-level language instructions~\cite{belkhale2024rthactionhierarchiesusing}, key-points~\cite{sundaresan2023kitekeypointconditionedpoliciessemantic, liu2024mokaopenworldroboticmanipulation}, or trajectories~\cite{niu2024llarvavisionactioninstructiontuning, gu2023rttrajectoryrobotictaskgeneralization, zawalski2024roboticcontrolembodiedchainofthought} to provide additional grounding to the robot policy. Despite the success of these methods, the generalization properties that these additional forms of grounding enable are still high-level in nature, and their benefits for simpler tasks may not necessarily transfer to more complex tasks -- ones that require further dexterity or object interactions. 

We hypothesize that the choice of mid-level representations is highly dependent on task-specific requirements. For instance, for a robot tasked with folding a shirt, a bounding box may help locate a shirt's general position but fails to provide actionable information on how to manipulate it. Similarly, depth-informed representations are crucial for contact-rich tasks such as handing over objects, but may be less important for overhead pick-and-place tasks. In order to give models stronger generalization capabilities for a wide variety of dexterous tasks, it is essential to explore representations that balance high-level abstraction and low-level actionable detail. These representations must not only encode spatial and geometric reasoning but also offer adaptability across diverse and dynamic environments.



In this paper, we find a set of mid-level representations that enhances the adaptability of a robot policy across a wide variety of environments. We first systematically study these representations across four critical dimensions: \textbf{object-centricity}, or understanding of the locations  and geometry of objects on the scene; \textbf{motion-centricity} or understanding of the future motion of the robot; \textbf{pose-awareness}, or understanding of spatial orientations; and \textbf{depth-awareness}, or understanding about three-dimensional structure and geometry. Through these experiments, we identify how different forms of spatial grounding align with task-specific requirements. We then propose a method to leverage these representations through a diffusion policy-based model conditioned on multiple experts outputting interpretable representations. We show that while different mid-level representations excel at different tasks, our method can leverage these task-specific benefits to achieve consistently higher performance on a wide range of environments. 

In addition, we further investigate how robot policies utilize the aforementioned representations. We find that reliance on structured signals presents a trade-off: policies that depend heavily on these representations can become more susceptible to overfitting and reduced robustness in noisy environments. To mitigate this, we introduce key architectural design choices that balance sensitivity to mid-level representations with resilience against spurious noise in these representations. Finally, we incorporate these representations as additional training signal.
We refer to the alignment of the demonstrations with these mid-level representation as \textit{self-consistency}, which can provide a weighting scheme for weighted imitation learning -- upweighting data with self-consistent representations. This approach further refines the policy’s ability to execute mid-level plans with greater precision and reliability.

Our policy, \MethodName, achieves a $24\%$ higher success rate than a standard diffusion policy baseline and an $11\%$ improvement over a baseline using language representations across a series of bimanual, dexterous tasks. Furthermore, our weighted imitation learning method enhances the policy's reliance on mid-level representations while preserving robustness to perturbations, resulting in a $10\%$ higher success rate compared to standard training. These results highlight that a deeper understanding of robot grounding can lead to significant improvements in robot success in dexterous, bimanual environments. 

\section{Related Works}
Training \emph{generalist} robot policies has been typically approached as a multi-task learning problem, where a single machine learning model is optimized for different behaviors and objectives. Many prior works have involved scaling up the breadth and diversity of robot data~\cite{dasari2020robonetlargescalemultirobotlearning, ebert2021bridgedataboostinggeneralization, khazatsky2024droidlargescaleinthewildrobot, liu2024rdt1bdiffusionfoundationmodel, zhao2024alohaunleashedsimplerecipe, kareer2024egomimicscalingimitationlearning, nasiriany2024robocasalargescalesimulationeveryday}. A key challenge with the multi-task policy learning regime is in obtaining policies that \emph{generalize} to new objects, task variants, environmental factors and so on. Towards this, a significant body of work has focused on generalizing robot policies to grasp new objects \cite{Saxena2006RoboticGO, mahler2017dexnet20deeplearning, james2018taskembeddedcontrolnetworksfewshot}. In this work, we are interested in achieving a similar form of generalization for dexterous manipulation~\cite{chen2022humanlevelbimanualdexterousmanipulation, zhao2023learningfinegrainedbimanualmanipulation}. Prior works in learning multi-task dexterous policies have struggled with generalizing to entirely new objects due to the high-dimensional observation and action spaces.

While the typical recipe to obtaining generalizable policies is to scale robot data, collecting such data remains prohibitively expensive. A promising alternative is to introduce structure into the end-to-end pixels-to-actions mapping. To provide robot models with a greater understanding of commonalities in robot tasks, planning, and behavior, several previous works have considered conditioning robot policies with higher-level representations of robot behavior~\cite{Shankar-2020-126755, pmlr-v100-lynch20a, hakhamaneshi2022hierarchicalfewshotimitationskill, wang2017robustimitationdiversebehaviors}. To increase the interpretability and controllability of robot policies, many works have instead conditioned on \textit{explicit} representations. These representations typically been specified either through goal images~\cite{nair2018visualreinforcementlearningimagined, ding2020goalconditionedimitationlearning}, video demonstrations~\cite{sontakke2023roboclipdemonstrationlearnrobot, du2023learninguniversalpoliciestextguided} or language \cite{goyal2021zeroshottaskadaptationusing, jang2022bczzeroshottaskgeneralization, zhou2024bridginglanguageactionsurvey}. While many earlier works have used higher-level conditioning information at the task specification level, recent works have works towards getting robots to achieve more specific goals, often specified in language \cite{huang2022languagemodelszeroshotplanners, ahn2022icanisay, huang2022innermonologueembodiedreasoning, black2023zeroshotroboticmanipulationpretrained}. 

One potential drawback to the hierachical learning framework is its rigidity in structure. Recently, many works have started viewing adding structure between robot perception and action as adding a more general "grounding" to the policies. A modern instantiation of this class of methods has been distilling additional knowledge into robot policies in the form of auxiliary tasks. For instance, some methods pre-train robot policies using regularization tasks such as visual question-answering (VQA) \cite{driess2023palmeembodiedmultimodallanguage, brohan2023rt2visionlanguageactionmodelstransfer, embodimentcollaboration2024openxembodimentroboticlearning, kim2024openvlaopensourcevisionlanguageactionmodel}, language planning \cite{sharma2022skillinductionplanninglatent, liang2023codepolicieslanguagemodel, ha2023scalingdistillingdownlanguageguided}, or spatial reasoning \cite{chen2024spatialvlmendowingvisionlanguagemodels}. Other approaches explicitly condition on higher-level representations, including language \cite{belkhale2024rthactionhierarchiesusing, zawalski2024roboticcontrolembodiedchainofthought} or image annotations \cite{gu2023rttrajectoryrobotictaskgeneralization, niu2024llarvavisionactioninstructiontuning, liu2024mokaopenworldroboticmanipulation}. However, many of these representations are disconnected from the physics of a robot's interaction with the world, and fail to capture the precise spatial and contextual details required for dexterous manipulation. For example, providing a language caption or object bounding box for a robot's workspace is not particularly informative about the object's geometry, orientation, affordances, and so on. In this paper, we investigate spatial mid-level representations that bridge the gap between high-level inputs (e.g., language commands or simple object markers) and the low-level action space of a robot. By grounding policies in richer spatial details, we aim to achieve better generalization and more reliable performance across a wide range of environments and tasks.

\section{Spatially-Grounded Mid-level Representations}
Robots that effectively generalize across diverse environments require an understanding of broad, high-level abstractions intrinsic to the real world, such as object geometry, spatial relationships, and motion dynamics. While one can hope to learn these relationships directly from end-to-end data, current large-scale robot policies that try to scale up imitation learning still struggle with performing dexterous tasks in environments that involve slight shifts in these properties. To address this issue, instead of implicitly relying on black-box feature extraction, we propose to explicitly incorporate representations that capture these mid-level abstractions, enabling robot policies to adapt more robustly to variations in real-world settings.

Concretely, we assume access to a dataset $\mathcal{D} = \{\tau_{1}, \tau_{2}, \ldots, \tau_{n}\}$, where each trajectory $\tau$ is a series of states and actions $(s_{1}, a_{1}), (s_{2}, a_{2}), \ldots, (s_{t}, a_{t})$. While typical end-to-end imitation learning aims at finding a mapping $\pi(a|s)$, we instead add a set experts that generate mid-level representations $\{E_{i}(s)\}_{i=1}^{k}$, each of which provides a specific type of grounding. We then aim to learn a policy $\pi(a|E_{1}(s), E_{2}(s), \ldots, E_{k}(s), s)$ which can leverage these representations to perform more robustly across diverse scenarios.

What representations would lead to the best performance for robot policies? There exists a hierarchy of representations, spanning from low-level geometric features to high-level symbolic structures such as language-based subtasks. For instance, language subtasks provide flexible, interpretable scaffolding that allows policies to be decomposed into modular instructions, facilitating transfer and reuse across different tasks. In parallel, higher granularity representations—like identified grasp locations—provide the precision required for dexterous manipulation and robust adaptation to slight variations in object geometry or environment layout. In this work, we concentrate on representations that hold the potential to enhance both the versatility and generality with which robots can perform dexterous tasks. By focusing on mid-level abstractions that capture critical factors—such as object geometry, spatial relationships, and motion dynamics—our goal is to equip robots with the adaptability needed to handle subtle variations in real-world environments more robustly.

We first investigate the utility of representations on four main axes: object-centricity, motion centricity, depth-awareness, and pose-awareness to provide a comprehensive spatial grounding for robot policies. In addition,  

\begin{enumerate}
    \item \textbf{Object-Centric Representations:} These experts focus on extracting information pertinent to each object in the scene, such as object poses, sizes, shapes, and potential interaction points (e.g., grasp points, keypoints for alignment). 

    \item \textbf{Motion-Centric Representations:} These experts capture and interpret the dynamic aspects of the environment by focusing on how objects, the robot, or other agents move and interact over time. For instance, they may encode object velocities, accelerations, potential collision points, or kinematic constraints. 
    
    \item \textbf{Depth-Aware Representations:} These experts leverage depth information to infer spatial relationships and physical constraints in the environment. For example, they can identify occlusions, measure distances between objects, and compute volumetric properties. By incorporating depth-aware reasoning, we enable robots to make more accurate and reliable decisions in 3D environments.

    \item \textbf{Pose-Aware Representations:} These experts encode the relative and absolute poses of objects, as well as the robot’s own pose, in a way that supports precise manipulation tasks. For example, pose-aware experts can compute alignment requirements for assembly tasks or predict optimal configurations for stable grasps. 
\end{enumerate}

These four axes—object-centric, motion-centric, depth-aware, and pose-aware—work together to provide a comprehensive spatial grounding for robot policies.

\begin{figure}
    \centering
    \includegraphics[width=0.5\textwidth]{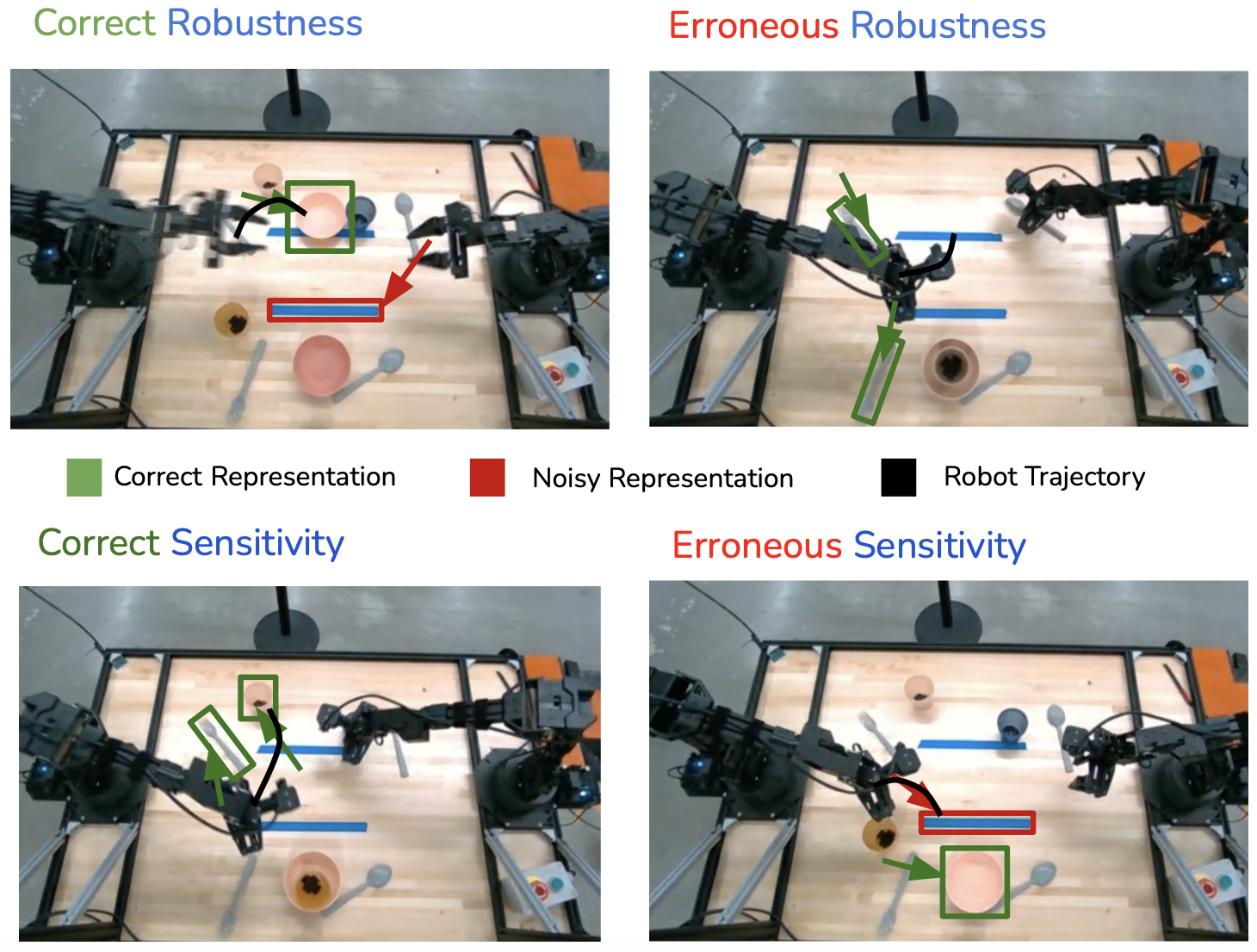}
    \caption{\textbf{The sensitivity-robustness tradeoff.} Policies need to follow their mid-level representations while being robust in erroneous noise to these representations.}
    \label{fig:sensitivity-robustness}
\end{figure}

\section{The Sensitivity-Robustness Tradeoff}
Previous works have predominantly focused on how various forms of grounding can enhance the performance of robotic policies. However, we argue that investigating mid-level relationships—specifically, how policies adhere to and utilize their representations—is equally crucial. By analyzing this relationship, we can view the mid-level representations as a \textit{bridge} between the sensory inputs of the policy and the lower-level joint actions. This enhances the interpretability of our policy by disentangling errors with the policy's mid-level decision making processing with its lower-level action generation. Through exploring this relationship, we identify a fundamental tradeoff between the sensitivity with which a robot follows its representations and its robustness to errors in these representations. 

Consider a household robot tasked with cleaning kitchenware, which receives mid-level representations from $k$ different experts $E_{1}, E_{2}, \ldots, E_{k}$ (see Figure~\ref{fig:sensitivity-robustness}). Suppose $E_{1}(s)$ provides the locations of objects of interest in the scene. A policy which utilizes these representations must consistently use these locations to output actions that move towards the correct target. The degree to which the policy follows this representation can be understood as \textit{sensitivity}. If one of the representations has an error, such as $E_{1}$ incorrectly detecting an object, the robot may attempt to interact with the wrong item, leading to improper handling or placement of kitchenware. The degree to which the policy is able to take reasonable actions in the prescence of these perturbations can be understood as \textit{robustness}.

This sensitivity-robustness tradeoff underscores the necessity of developing robot policies that balance adherence to mid-level representations with the ability to remain adaptable and resilient in the face of environmental variations. This balance ensures that, while policies leverage detailed, task-relevant information for precise manipulation, they do not become overly dependent on specific features that may change or vary in different contexts. 

Our usage of \textit{spatially-aware} mid-level representations allows us to directly quantify the tradeoff between generalization and executability in robotic policies. We propose two key metrics to measure this tradeoff:

\begin{enumerate}
    \item \textbf{Sensitivity Score (SS):}  
    The Sensitivity Score quantifies the extent to which the policy adheres to the provided mid-level representations during task execution. Specifically, it measures how variations or perturbations in the representations influence the resulting trajectories of the robot. Formally, let $f(s, E, \tau)$ represent a function that evaluates the adherence of the trajectory \( \tau \) to the representations $ E = \{E_{1}(s), E_{2}(s), \ldots, E_{k}(s)\} $ given the state $s$. The Sensitivity Score is defined as:
    $$\text{SS}(E) = \mathbb{E}_{s, \tau} \left[ \text{Adherence}(E(s), \tau) \right]$$
    where $\text{Adherence}(\cdot)$ is a metric quantifying the alignment between the policy's trajectory and a particular mid-level representation. A lower SS indicates that the policy closely follows the representations. Conversely, a higher SS implies that the policy is less reliant on specific representations. Further information on how $\text{Adherence}(\cdot)$ is computed for each representation can be found in Appendix~\ref{appendix:metrics}.
   
\item \textbf{Robustness Index (RI):}  
The Robustness Index measures the policy's ability to maintain stable and effective performance when perturbations are introduced to the mid-level representations. For each state \( s \) in a trajectory \( \tau \), and for each mid-level representation \( E_i(s) \) generated by the experts, we apply a Gaussian perturbation:
\[
\tilde{E}_i(s) = E_i(s) + \epsilon_i, \quad \epsilon_i \sim \mathcal{N}(0, \sigma^2)
\]
The perturbed representations \( \tilde{E}_i(s) \) are then used as inputs to the policy:
\[
\pi(a \mid \tilde{E}_1(s), \tilde{E}_2(s), \ldots, \tilde{E}_k(s), s)
\]

Let \( P_j^{(\text{perturb})}(\sigma) \) represent the policy's performance under the \( j \)-th Gaussian perturbation with standard deviation \( \sigma \), and \( P_0 \) its performance with the original, unperturbed representations. The Robustness Index is defined as the mean of the performance ratios across all perturbations:
\[
\text{RI}(\sigma) = \frac{1}{m} \sum_{j=1}^{m} \frac{P_j^{(\text{perturb})}(\sigma)}{P_0}
\]
where \( m \) is the number of distinct Gaussian noise realizations applied to the representations. A higher RI signifies that the policy remains resilient and maintains its effectiveness despite perturbations in the representations. Conversely, a lower RI suggests that the policy is prone to performance degradation when the representations are altered, highlighting potential overfitting to the exact representations provided during training.

\end{enumerate}

\section{Architecture}
\begin{figure*}[htbp]
  \centering
  \includegraphics[width=0.95\textwidth]{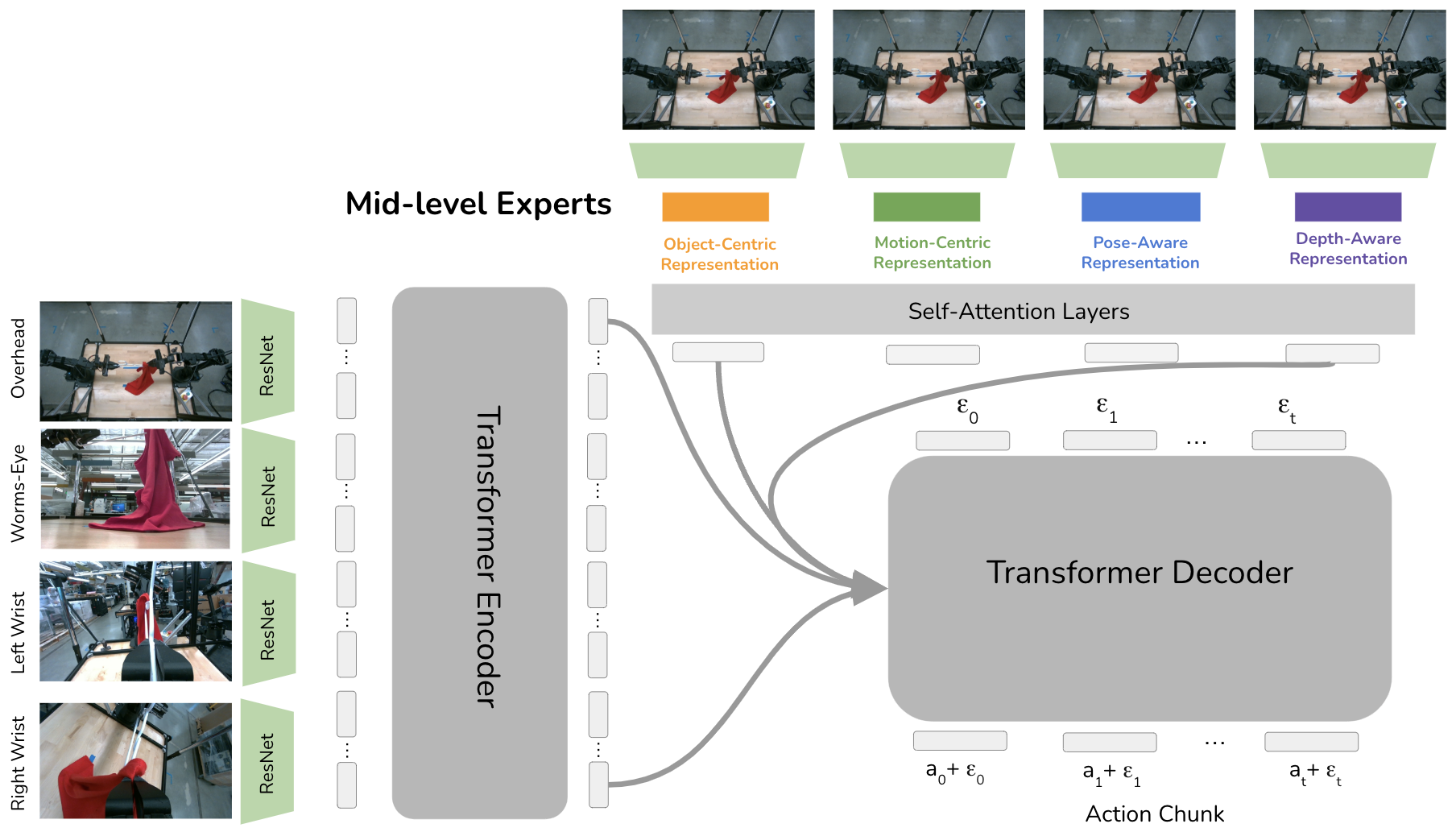}
  \caption{\textbf{Policy Architecture.} Four images are passed into a transformer encoder. In addition, an image is fed into each individual mid-level expert. The results embeddings are passed into the transformer decoder through cross-attention.}
  \label{fig:architecture}
\end{figure*}

We implement our method on a diffusion policy similar to the one proposed in \cite{zhao2024alohaunleashedsimplerecipe}.  The policy takes as input $4$ images from different viewpoints ($2$ third-person images and $2$ wrist images) and directly outputs $12$ absolute joint positions--$6$ for each arm--as well as a continuous gripper value per each gripper. Each image is fed through a separate ResNet50 encoder, before being processed with a transformer encoder to obtain image embeddings. At each state, we denoise the decoder predicts $t = 10$ action chunks simultaneously with a transformer. See Figure~\ref{fig:architecture} for a depiction of our architecture.

\subsection{Mid-level Experts}
We design our architecture motivated by the sensitivity-robustness tradeoff. At each state, the robot must discern which representations are pertinent for the task at hand and output actions which follow these representations. Meanwhile, if any of the representations have noise, the policy must output reasonable actions that maintain task performance despite these disturbances. To achieve this, we employ three key design choices: 

\begin{enumerate}
    \item \textbf{Diverse Mid-level Experts} We employ $k = 4$ mid-level experts to output representations corresponding to our aforementioned axes of grounding. In particular, we have separate object-centric, motion-centric, pose-aware, and depth-aware representations. These correspond to bounding boxes, trajectory traces, grasp plans, and depth-aware traces. 
    
    \item \textbf{Attention-based Mid-level Gating:} In order to combine our experts, we use multi-headed attention to provide an early form of gating. Specifically, the embeddings are processed into a multidimensional tensor $z = \text{MultiHead}(R_1, R_2, R_3, R_4)$, where each \( R_i \) is the representation from expert \( E_i \). This gating strategy dynamically weights each expert's contribution based on the current state, enabling effective integration of diverse representations.
    
    \item \textbf{Cross-Attention Mechanism:} We then perform cross-attention between the processed mid-level embeddings $z$ and the image embeddings. This allows the policy to further use information from its images to determine which representations to determine which representations to emphasize for action selection. This ensures that the policy dynamically prioritizes the most relevant representations based on the current visual and contextual information, as well as be more robust to errors in these representations. 
\end{enumerate}

\subsection{Training}
To train our policy, we adopt a two-stage approach. In the first stage, the expert mid-level representations are trained separately on data tailored to each representation. This data is purely relabeled from demonstrations and proprioception.
\\
For motion-centric representations, we use proprioception data, which refers to the sensor data related to the arm's position and movement, to process the arm's trajectories at future points in time. We utilize a trajectory length of 10 points, with one annotation every 10 timesteps. We project these representations into the observation frame using the camera intrinsics and extrinsics. In simulation, these values are provided by the simulator, while in the real world, they are calibrated using an AprilTag board. 
\\
\indent We base our object-centric representations on an off-the-shelf OWL-ViT (153M params). The language prompts are carefully tuned for each task to minimize noise as much as possible. For pose-aware representations on top of bounding boxes, we combine proprioception by processing states where the arm comes into contact with objects. We then retrain ResNet34 models (21M params) on top of this relabelled data. We find that that for more dexterous tasks, this is important in order to avoid significant decreases in the frequency of robot control by using an OWL-ViT in-the-loop.  \\
\indent Once the expert modules are trained independently, their parameters are frozen. Then, the policy network trained end-to-end with a noise prediction loss. During inference time, each of the expert models are executed asynchronously.

\subsection{Self-Consistency}
\begin{figure}
    \centering
    \includegraphics[width=0.5\textwidth]{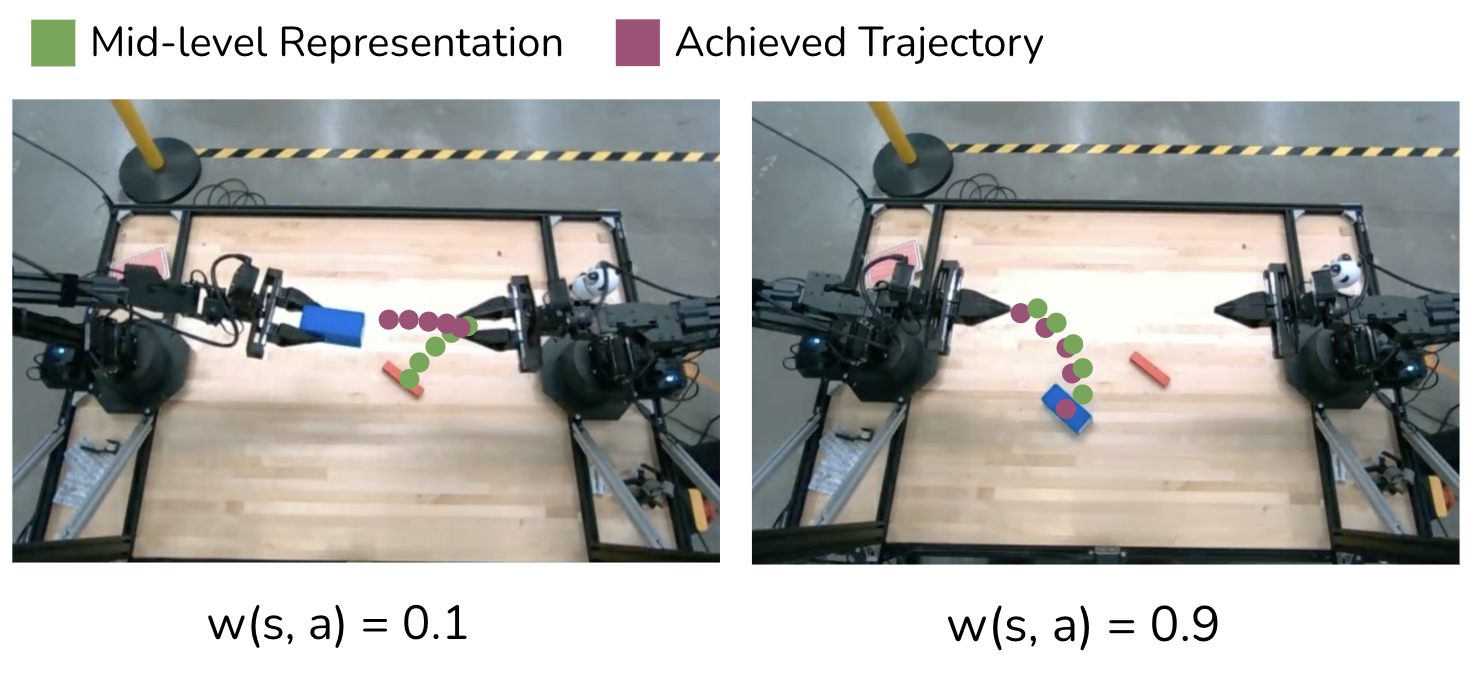}
    \caption{\textbf{Self-Consistency.} On the left image, the robot's achieved trajectory doesn't match its mid-level representation, which leads to a lower weight. On the right, the robot follows its representation, leading to a higher weight.}
    \label{fig:self-consistency}
\end{figure}

A primary challenge in integrating mid-level representations into robot policies lies in ensuring that the policy consistently follows the guidance provided by these representations. Object-centric attributes, depth-aware insights, or pose-aware signals serve as mid-level “expert outputs” that the policy is incentivized to replicate. However, direct supervision through standard behavioral cloning (BC) can lead to inconsistencies, especially when the mid-level predictions are noisy or only partially correct. Such inconsistencies ultimately degrade the policy’s ability to effectively utilize the expert signals.

Our proposed method is analogous to reinforcement learning with a hand-crafted advantage function. In RL, the loss function typically incorporates an advantage term, given by:
\[
\mathcal{L}_{\text{RL}} = \mathbb{E}_{(s, a) \sim \mathcal{D}} \left[ A(s, a) \cdot \mathcal{L}_{\text{PG}} \left( \pi(a \mid s) \right) \right]
\]
where $A(s, a)$ represents the advantage function, which modulates the policy gradient loss $\mathcal{L}_{\text{PG}}$ based on the estimated benefit of selecting action $a$ in state $s$. Similarly, our approach integrates mid-level expert outputs as implicit guidance in scenarios where no explicit reward signal is available. Instead of an advantage function, we introduce \emph{self-consistency weights} \( w(x) \), which serve to emphasize reliable expert guidance. The corresponding loss function is:
\[
\mathcal{L}_{\text{policy}} = \mathbb{E}_{(x, a) \sim \mathcal{D}} \left[ w(x) \cdot \mathcal{L}_{\text{BC}} \left( \pi(a \mid x), a \right) \right]
\]
where $w(x)$ reflects how well the mid-level expert outputs align with the ground truth or improve task success. Notably, our representations encode the policy's desired future behavior, similar to how an advantage function models a policy's expected future reward. By structuring policy learning in this way, our method ensures that mid-level expert outputs provide meaningful guidance, akin to an advantage function in reinforcement learning. This approach prioritizes high-quality supervision, improving the policy's ability to effectively utilize expert-generated representations. By iteratively refining the training data and adjusting the weighting of consistent samples, our method creates a feedback loop that promotes tighter self-consistency between policy actions and mid-level expert outputs. See Algorithm~\ref{alg:self-consistency} for more details. The exact method to compute the weights can be viewed in Appendix~\ref{appendix:weights}.

\begin{algorithm}[t]
\caption{Weighted Self-Consistency Training}
\label{alg:self-consistency}
\begin{algorithmic}[1]
\REQUIRE Policy $\pi_\theta$ (with parameters $\theta$), Mid-level experts $f_{\text{expert}}$,\\ 
Training dataset $D=\{(s_i,a_i)\}$, Learning rate $\eta$
\vspace{5pt}
\STATE Initialize policy parameters $\theta$
\REPEAT
  \STATE Sample a minibatch of $B$ states/actions $\{(s_i,a_i)\}_{i=1}^B$ from $D$
  \STATE Compute mid-level outputs $m_i = f_{\text{expert}}(s_i)$ for each sample
  \STATE Compute self-consistency weights: \\
  \[
    w_i = \exp\left( -\frac{1}{|\mathcal{E}|} \sum_{E \in \mathcal{E}} \lambda_E \cdot \text{Adherence}(E(s_i), \tau) \right)
  \]
  \STATE Compute weighted BC loss: 
  \[
    \mathcal{L}_{\text{BC}} = \frac{1}{B}\sum_{i=1}^B w_i \,\ell\bigl(\pi_\theta(s_i), a_i\bigr)
  \]
  \STATE Update policy parameters: $\theta \leftarrow \theta - \eta \,\nabla_\theta \,\mathcal{L}_{\text{BC}}$
\UNTIL{convergence}
\end{algorithmic}
\end{algorithm}

\section{Experiments}

Our goal is to evaluate the effectiveness of mid-level representations as grounding for training robotic policies, focusing on their ability to improve task performance across diverse scenarios. Specifically, we aim to assess how different forms of spatial grounding—object/motion-centricity, pose-awareness, and depth-awareness—impact a robot's ability to generalize, execute precise actions, and recover from noisy inputs.

\begin{enumerate}
\item Are mid-level representations effective in improving policy generalization performance across a range of tasks and environments?
\item What types of tasks benefit the most from specific mid-level representations, such as object/motion-centricity, pose-awareness, or depth-awareness, and how do these align with task-specific requirements?
\item Can a policy effectively utilize multiple sources of mid-level representations, and how does the integration of these diverse signals impact task performance and generalization?
\item Which policy architecture offers the best tradeoff between responsiveness to structured mid-level representations and robustness to noise or spurious inputs?
\item Can mid-level representations be effectively used as supervision signals during training to enhance policy precision and generalization across tasks?
\end{enumerate}

\subsection{Simulation Environment} 
\begin{figure}
    \centering
    \includegraphics[width=0.5\textwidth]{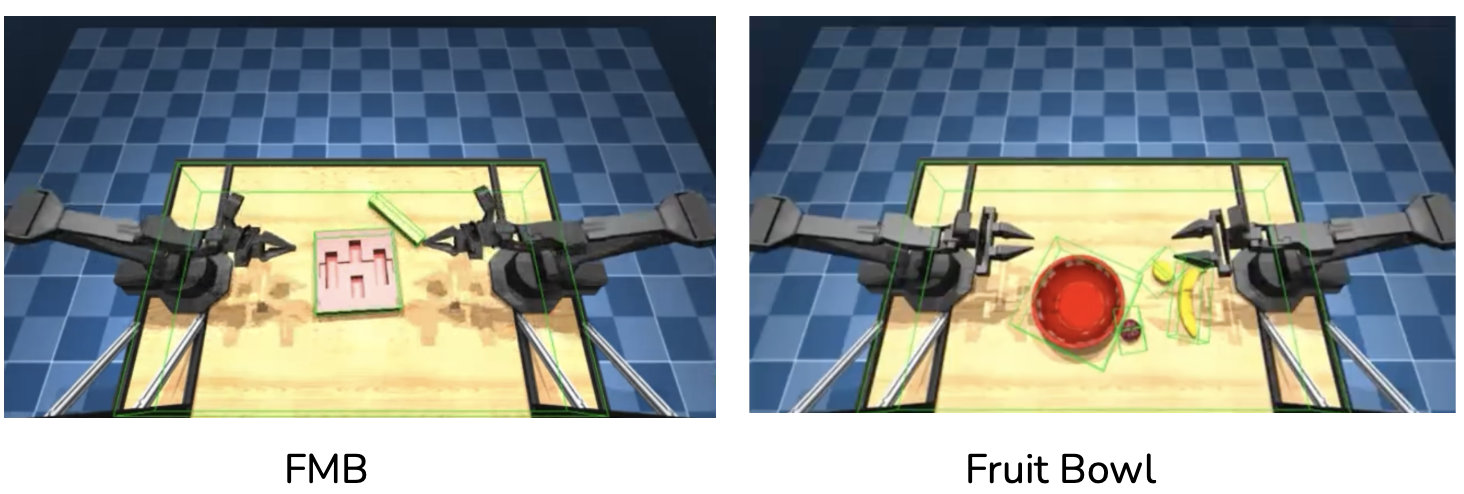}
    \caption{\textbf{Simulation Tasks.} }
    \label{fig:sim-tasks}
\end{figure}

We evaluate our method on the Aloha Unleashed simulation environment \cite{zhao2024alohaunleashedsimplerecipe}. This environment consistent of a bimanual parallel-jaw gripper robot workcell. The observations contain images from 4 different points of view: the overhead camera, worms-eye camera (facing the robot), and two wrist cameras. 

\begin{itemize}
    \item \textbf{Single Insertion:} The robot must pick up a peg with one arm, and a block containing a hole with another. Then, it must align the peg with the block and insert it into the hole.
    
    \item \textbf{FMB Assembly:} The robot must pick up a multiple blocks and place it into its appropriate slots. Each of the blocks must be placed in precise locations that require the robot to have good understanding of object geometry.
    
    \item \textbf{Fruit Bowl:} The robot must arrange a variety of fruits into a bowl, ensuring efficient use of space or adherence to a specific pattern. This task tests the robot's ability to handle objects of different shapes and sizes while reasoning about spatial configurations.

    \item \textbf{Pen Handover:} The task requires the robot to perform a smooth and reliable pen handover to a human. This includes grasping the pen, orienting it correctly for comfortable use, and transferring it to the human's hand without dropping or misalignment. The task tests precision grip, human-robot interaction, and timing coordination.
\end{itemize}

\begin{figure*}[ht]
    \centering
    \includegraphics[width=\textwidth]{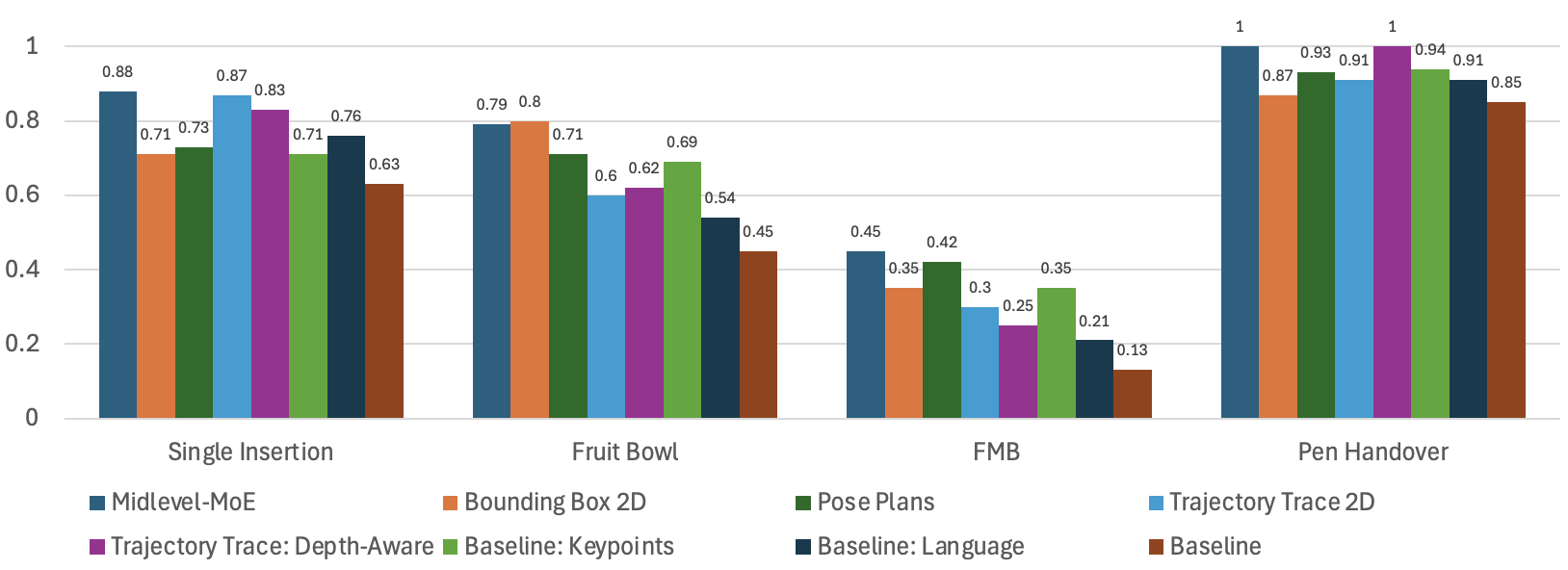}
    \caption{\textbf{Simulation Results.} Mid-level MoE achieves a $24\%$ higher success rate over a standard diffusion policy baseline. It performs consistently well over different tasks by leveraging different representations.}
    \label{fig:sim_results}
\end{figure*}
\subsection{Real-World Environment}
We evaluate our method on a similar real-world environment consisting of two 6-DoF ViperX robot arms with parallel-jaw grippers. The observations perspectives are the same as in simulation.

\begin{itemize}
    \item \textbf{Kitchen Stack:} The robot must organize and stack a set of kitchen items, such as bowls, plates, and cups, in an orderly manner on a designated shelf or surface. The task emphasizes spatial reasoning, stability prediction, and careful object placement.

    \item \textbf{Cup Stacking:}  The robot must pick up multiple cups on the scene and stack the cups. This tasks tests the ability of the robot to generalize across different combinatorial object locations across the workstation.
    
    \item \textbf{Shirt Hanging:}  The task requires the robot to hang a shirt on a hanger. The steps include flattening the shirt, picking up a hanger from a rack, moving the shirt to a desirable location, placing the hangar onto the shirt, then picking up the hangar and putting it back onto the rack.

    \item \textbf{Shoelace Tying:} The robot must tie the laces of a shoe into a bow. This involves centering the shoe, straightening the laces, crossing them over, forming loops, and tightening the bow. The task emphasizes a robot's fine motor control.
\end{itemize}

\subsection{Experiment Setup}
To evaluate the impact of a singular representation on our policy, we ablate our method by conditioning on a single expert. In addition, we provide two ablations based on prior works investigating a single representation: a keypoints-based ablation based on MOKA~\cite{liu2024mokaopenworldroboticmanipulation} and a language baseline based on RT-H~\cite{belkhale2024rthactionhierarchiesusing}. In the keypoint ablation, we identify important points of interest in the image by querying a VLM. For RT-H, we relabel robot demonstrations with the language "move the arm left/right/up/down." For each environment in simulation and the real-world, we vary the object locations, add distractor objects, and change the lighting conditions.

\section{Analysis}
\begin{figure*}[ht]
    \centering
    \includegraphics[width=\textwidth]{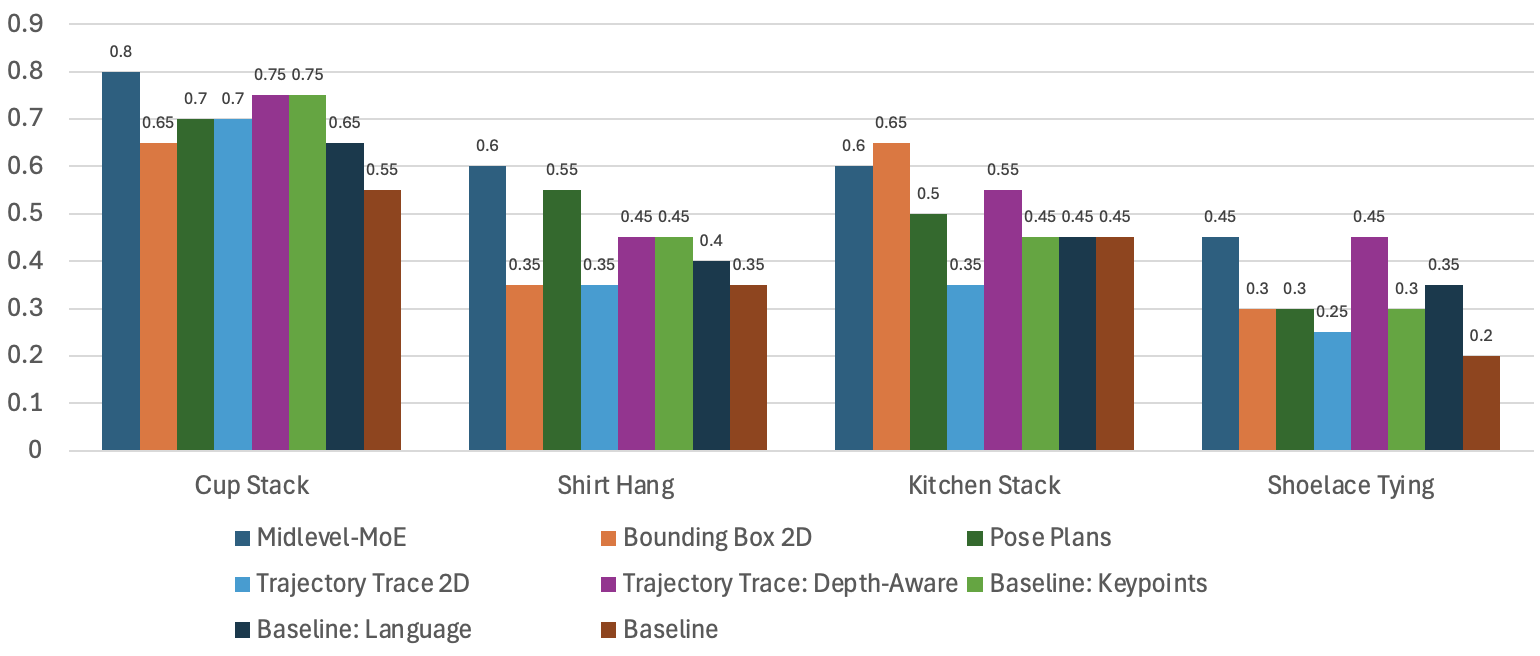}
    \caption{\textbf{Real-World Results.} There are clear differences in the benefits that different representations provide for tasks in the real world.}
    \label{fig:real_results}
\end{figure*}
 
\subsection{Mid-level Representations have \textit{Task-Specific} Benefits}
Figures~\ref{fig:sim_results} and \ref{fig:real_results} present our experimental results in both simulation and real-world environments. We find that while all representation ablations outperform the baseline with no representations, their effectiveness varies depending on the task. More specifically, we observe that \textbf{motion-centric} representations (trajectory trace) achieve higher success rates in \textit{Single Insertion} and \textit{Cup Stack}, \textbf{object-centric} representations (bounding box) perform better in \textit{Fruit Bowl} and \textit{Kitchen Stack}, \textbf{pose-aware} representations (pose plans) excel in \textit{FMB} and \textit{Shirt Hang}, and \textbf{depth-aware} representations (trajectory trace: depth-aware) yield the highest success in \textit{Pen Handover} and \textit{Shoelace Tying}. These findings are consistent with our qualitative understanding of the grounding required for each task. For instance, tasks like \textit{Fruit Bowl} and \textit{Kitchen Stack} involve rearranging multiple objects within a cluttered scene, making object-centric representations particularly effective. In contrast, \textit{Pen Handover} and \textit{Shoelace Tying} necessitate a precise understanding of object relationships in 3D space, emphasizing the importance of depth-aware representations for accurate spatial reasoning and fine-grained manipulation.

\begin{table}[!htb]
    \centering
    \begin{tabular}{l|c|c}
    \toprule
    \textbf{Representation} & \textbf{Single Insertion} &\textbf{FMB}\\
    \midrule
    Bounding Box 2D & 0.85 & 0.51 \\
    Grasp Plans & 0.87 & 0.67 \\
    Trajectory Trace 2D & 0.92 & 0.45 \\
    Trajectory Trace: Depth-Aware & 0.97 & 0.57  \\
    \bottomrule
    \end{tabular}
    \caption{\textbf{Performance Across Representations with Ground Truth.} Even with ground-truth mid-level representation, the success rates policies with different mid-level experts differ by at least $12\%$}
    \label{table:ground_truth_representations}
\end{table}

The advantages of different mid-level representations become more evident when analyzing their impact with ground-truth data depicted in Table~\ref{table:ground_truth_representations}. For \textit{Bounding Box 2D} and \textit{Grasp Plans}, values are directly derived from the simulation, while for \textit{Trajectory Trace 2D} and \textit{Trajectory Trace: Depth-Aware}, future trajectory values are estimated based on the robot arm's current pose and the pose of the object it interacts with. Since trajectory estimation becomes more complex when multiple objects are involved, we evaluate performance in two representative tasks: \textit{Single Insertion} and \textit{FMB}. Our results show that in \textit{Single Insertion}, motion-centric representations, such as trajectory traces, outperform object-centric representations ($94\%$ average success rate vs $85\%$. Meanwhile, in \textit{FMB}, pose-aware representations like \textit{Grasp Plans} yield significantly better performance $67\%$ success rate. This highlights the importance of selecting representations that align with task-specific requirements.

\subsection{\MethodName~can effectively utilize multiple different representations to generalize across a broad range of tasks.}
Figures~\ref{fig:sim_results} and \ref{fig:real_results} also compare \MethodName~to a language baseline with lower-level language commands similar to \cite{belkhale2024rthactionhierarchiesusing}, a keypoints baseline with object-centric points of interest similar to ~\cite{liu2024mokaopenworldroboticmanipulation}, and a simple diffusion policy baseline with no mid-level representations. Our method has an average of $69.6\%$ success over all tasks compared to $45.1\%$ success rate for the no-representation baseline, $51.5\%$ success rate for the language baseline, and $58\%$ success rate for the keypoints baseline. This highlights the usefulness of having more granular spatial grounding compared to higher-level representations like language for more dexterous tasks. 

In addition, our results indicate that \MethodName~can leverage the task-specific benefits of different representations across different tasks. While \MethodName~doesn't always perform better than the best representation for a particular task (such as compared to bounding boxes in the \textit{Kitchen Stack} task), it consistently performs at a high across all evaluated tasks, Likewise, although we find some tasks that the keypoints and language baselines do well on (\textit{Cup Stack} and \textit{Single Insertion} respectively), since different tasks require different grounding, they don't consistently lead to large increases in policy performance.

\begin{figure*}[ht]
    \centering
    \includegraphics[width=\textwidth]{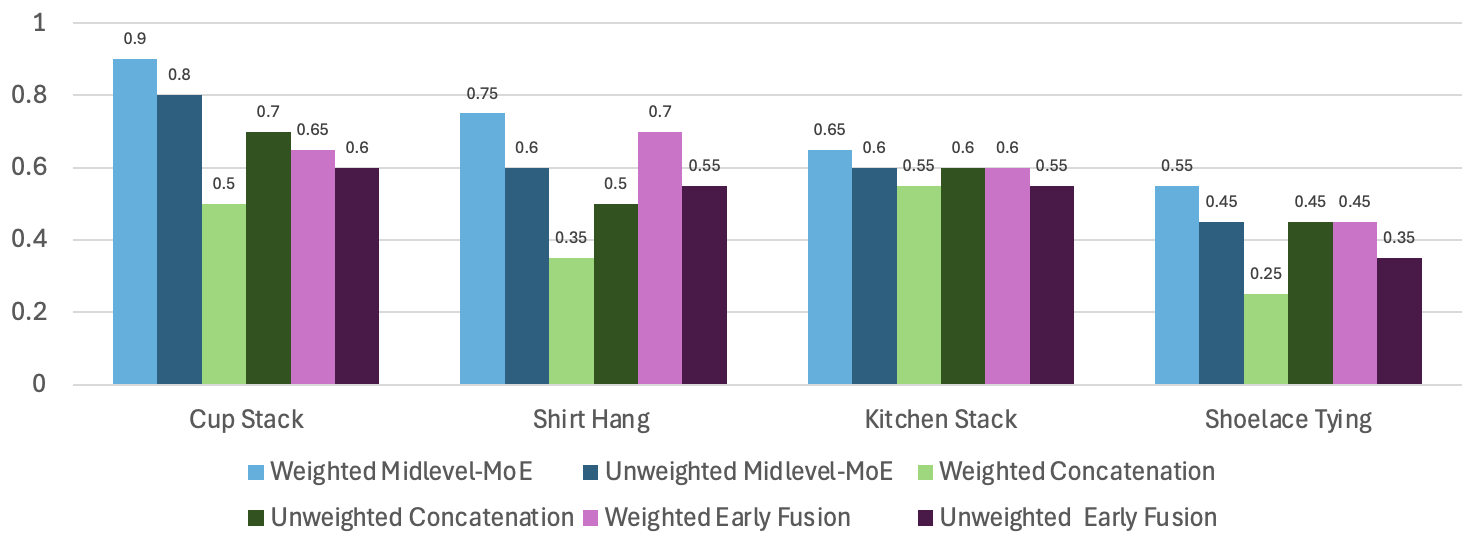}
    \caption{\textbf{Architecture and Self-Consistency Ablation.} Weighted \MethodName~achieves an average of $10\%$ higher success rate than unweighted over $4$ real-world tasks.}
    \label{fig:real_results}
\end{figure*}

\begin{table*}[!htb]
    \centering
    \begin{tabular}{lccccc}
    \toprule
    \textbf{Method} & \textbf{SS: Bounding Box} &  \textbf{SS: Grasp Plan} & \textbf{SS: Trace 2D} & \textbf{SS: Trace Depth-Aware} & \textbf{RI} \\
    \midrule
    Weighted \MethodName & 0.06 & 0.21 & 0.12 & 0.20 & 0.80 \\
    Unweighted \MethodName & 0.1 & 0.36 & 0.16 & 0.25 & 0.86 \\
    Weighted Concatenation & 0.10 & 0.24 & 0.17 & 0.23 & 0.51 \\
    Unweighted Concatenation & 0.15 & 0.41 & 0.20 & 0.27 & 0.75 \\
    Weighted Early Fusion & 0.14 & 0.30 & 0.20 & 0.27 & 0.73 \\
    Unweighted Early Fusion & 0.19 & 0.47 & 0.25 & 0.35 & 0.80 \\
    \bottomrule
    \end{tabular}
    \caption{\textbf{Sensitivity and Robustness.} Comparison of average sensitivity scores and robustness indices across various architectures. Policies with lower sensitivtiy scores and higher robustness indices tend to perform better.}
    \label{table:sensitivity}
\end{table*}

\begin{table}[!htb]
    \centering
    \begin{tabular}{l|c|c|c}
    \toprule
    \textbf{Method} & \textbf{\MethodName} & \textbf{Concatenation} & \textbf{Early Fusion}  \\
    \midrule
    Simulation & 0.78 & 0.72 & 0.65  \\
    Real & 0.61 & 0.56 & 0.52 \\
    \bottomrule
    \end{tabular}
    \caption{\textbf{Architecture Ablation } Average success rates for different architectures overs simulation and real tasks.}
    \label{table:architecture}
\end{table}

\subsection{Different Architectures offer Different Tradeoffs between Sensitivity and Robustness}
We ablate different policy architectures in Tables ~\ref{table:architecture} and ~\ref{table:sensitivity}. Table ~\ref{table:architecture} records the average success rates for our architecture across all simulation and real-world tasks. Meanwhile, Table~\ref{table:architecture} records the sensitivity scores for each of our mid-level experts as well as the robustness index. The robustness index is computed by adding Gaussian noise with standard deviation $0.1$ to all of the representations and finding the ratio between the success rate with and without perturbation.

In particular, we find that our method outperforms early fusion and concatenation by an average of $11\%$ and $7.5\%$, respectively. Notably, these performance enhancements are strongly correlated with the sensitivity scores observed across different architectures. \MethodName~achieves an average sensitivity score of $0.2175$, which is significantly lower than those of concatenation ($0.2575$) and early fusion ($0.315$). This reduction in sensitivity suggests that \MethodName~is more adept at focusing on relevant mid-level representations, thereby enhancing policy performance and increasing success rates. Furthermore, our analysis of robustness indices reveals that concatenation methods exhibit lower robustness ($0.75$) compared to attention-based approaches, including \MethodName~($0.86$). This higher robustness indicates that attention-based methods are better at maintaining stable performance under varying conditions and potential perturbations. The combination of lower sensitivity and higher robustness in \MethodName~not only contributes to its superior average performance but also ensures more reliable and consistent outcomes across diverse tasks.

Overall, these findings underscore the effectiveness of \MethodName~in leveraging task-specific representations through attention mechanisms. By selectively attending to pertinent information and maintaining robustness, \MethodName~demonstrates a balanced approach that enhances both the efficiency and reliability of policy execution across multiple tasks.

\subsection{Self-Consistency Leads to Higher Sensitivity Scores}
Table~\ref{table:sensitivity} demonstrates that policies trained with self-consistency exhibit increased sensitivity scores, indicating an enhanced ability to effectively leverage relevant mid-level features. In contrast, employing the Weighted Imitation Learning algorithm described in Algorithm~\ref{alg:self-consistency} results in an average sensitivity score that is $25.5\%$ lower than that of unweighted approaches across the three architectures. This reduction suggests that weighted imitation learning enables policies to focus more precisely on pertinent features, potentially minimizing the influence of noisy or irrelevant information.

However, the robustness indices reveal a trade-off: weighted imitation learning also leads to a $9\%$ lower average robustness index. This decrease implies that while the policies become more selective in their feature utilization, they may be slightly less resilient to variations or adversarial perturbations in the input data. Balancing sensitivity and robustness is crucial, as excessive sensitivity can make models vulnerable to minor input changes, whereas reduced robustness might compromise performance in dynamic or unpredictable environments.

We find that policy architecture plays a crucial role in performance. Figure ~\ref{fig:self-consistency} shows results for various weighted and unweighted architectures across $4$ real-world tasks. Interestingly, Weighted \MethodName~has an average of $10\%$ higher performance than \MethodName~across the $4$ tasks.  This suggests that the benefits of more targeted feature utilization outweigh the slight decrease in robustness. However, self-consistency doesn't always improve performance. Notably, we see a decrease in performance for the weighted concatenated architecture versus unweighted (an average of $26\%$ drop in performance). A reasonable explanation for this is a significant lower robustness index for the concatenated architecture ($0.51$). While self-consistency does indeed lower the sensitivity score for \textit{Weighted Concatenation}, the policy struggles with being robust to noise and spurious correlations in these representations. 

\section{Conclusion} 
\label{sec:conclusion}
In this paper, we examine the types of grounding necessary for generalization in dexterous, bimanual robotic tasks. Our findings reveal that the effectiveness of different representations in facilitating generalization is inherently task-dependent. To address this challenge, we first identify four key representations—object-centric, motion-centric, pose-aware, and depth-aware—that achieve strong performance across a diverse set of tasks. We then introduce \MethodName, a novel policy that integrates multiple mid-level experts, each specializing in a distinct representation. Through our investigation, we uncover a fundamental tradeoff between sensitivity—the degree to which a policy adheres to a mid-level representation—and robustness—its resilience to noise within these representations. We demonstrate that applying weighted imitation learning to the \MethodName~architecture effectively reduces sensitivity while preserving a high level of robustness. Ultimately, this work takes a step toward a deeper understanding of the grounding strategies necessary for scaling robot policies to broader and more complex environments.

\smallskip \noindent \textbf{Limitations and Future Direction.} 
Our methodology comes with several limitations. First, due to frequency constraints in dexterous tasks, we distill our mid-level representations into smaller policies. Training policies to generate more generalist representations can introduce challenges related to inference speed. A promising future direction is to enable policies to condition on representations asynchronously while maintaining both high frequency and strong performance on individual tasks. In addition, our self-consistency method relies on hand-designed metrics for each task. While this is feasible for the tasks we have evaluated, scaling this approach to a broader range of tasks would require either extensive manual tuning or automated metric discovery methods. 

\newpage

\bibliographystyle{plainnat}
\bibliography{references}

\begin{thebibliography}{41}
\providecommand{\natexlab}[1]{#1}
\providecommand{\url}[1]{\texttt{#1}}
\expandafter\ifx\csname urlstyle\endcsname\relax
  \providecommand{\doi}[1]{doi: #1}\else
  \providecommand{\doi}{doi: \begingroup \urlstyle{rm}\Url}\fi

\bibitem[Ahn et~al.(2022)]{ahn2022icanisay}
Michael Ahn et~al.
\newblock Do as i can, not as i say: Grounding language in robotic affordances, 2022.
\newblock URL \url{https://arxiv.org/abs/2204.01691}.

\bibitem[Belkhale et~al.(2024)Belkhale, Ding, Xiao, Sermanet, Vuong, Tompson, Chebotar, Dwibedi, and Sadigh]{belkhale2024rthactionhierarchiesusing}
Suneel Belkhale, Tianli Ding, Ted Xiao, Pierre Sermanet, Quon Vuong, Jonathan Tompson, Yevgen Chebotar, Debidatta Dwibedi, and Dorsa Sadigh.
\newblock Rt-h: Action hierarchies using language, 2024.
\newblock URL \url{https://arxiv.org/abs/2403.01823}.

\bibitem[Black et~al.(2023)Black, Nakamoto, Atreya, Walke, Finn, Kumar, and Levine]{black2023zeroshotroboticmanipulationpretrained}
Kevin Black, Mitsuhiko Nakamoto, Pranav Atreya, Homer Walke, Chelsea Finn, Aviral Kumar, and Sergey Levine.
\newblock Zero-shot robotic manipulation with pretrained image-editing diffusion models, 2023.
\newblock URL \url{https://arxiv.org/abs/2310.10639}.

\bibitem[Brohan et~al.(2023)]{brohan2023rt2visionlanguageactionmodelstransfer}
Anthony Brohan et~al.
\newblock Rt-2: Vision-language-action models transfer web knowledge to robotic control, 2023.
\newblock URL \url{https://arxiv.org/abs/2307.15818}.

\bibitem[Chen et~al.(2024)Chen, Xu, Kirmani, Ichter, Driess, Florence, Sadigh, Guibas, and Xia]{chen2024spatialvlmendowingvisionlanguagemodels}
Boyuan Chen, Zhuo Xu, Sean Kirmani, Brian Ichter, Danny Driess, Pete Florence, Dorsa Sadigh, Leonidas Guibas, and Fei Xia.
\newblock Spatialvlm: Endowing vision-language models with spatial reasoning capabilities, 2024.
\newblock URL \url{https://arxiv.org/abs/2401.12168}.

\bibitem[Chen et~al.(2022)Chen, Wu, Wang, Feng, Jiang, McAleer, Geng, Dong, Lu, Zhu, and Yang]{chen2022humanlevelbimanualdexterousmanipulation}
Yuanpei Chen, Tianhao Wu, Shengjie Wang, Xidong Feng, Jiechuang Jiang, Stephen~Marcus McAleer, Yiran Geng, Hao Dong, Zongqing Lu, Song-Chun Zhu, and Yaodong Yang.
\newblock Towards human-level bimanual dexterous manipulation with reinforcement learning, 2022.
\newblock URL \url{https://arxiv.org/abs/2206.08686}.

\bibitem[Collaboration et~al.(2024)]{embodimentcollaboration2024openxembodimentroboticlearning}
Open X-Embodiment Collaboration et~al.
\newblock Open x-embodiment: Robotic learning datasets and rt-x models, 2024.
\newblock URL \url{https://arxiv.org/abs/2310.08864}.

\bibitem[Dasari et~al.(2020)Dasari, Ebert, Tian, Nair, Bucher, Schmeckpeper, Singh, Levine, and Finn]{dasari2020robonetlargescalemultirobotlearning}
Sudeep Dasari, Frederik Ebert, Stephen Tian, Suraj Nair, Bernadette Bucher, Karl Schmeckpeper, Siddharth Singh, Sergey Levine, and Chelsea Finn.
\newblock Robonet: Large-scale multi-robot learning, 2020.
\newblock URL \url{https://arxiv.org/abs/1910.11215}.

\bibitem[Ding et~al.(2020)Ding, Florensa, Phielipp, and Abbeel]{ding2020goalconditionedimitationlearning}
Yiming Ding, Carlos Florensa, Mariano Phielipp, and Pieter Abbeel.
\newblock Goal-conditioned imitation learning, 2020.
\newblock URL \url{https://arxiv.org/abs/1906.05838}.

\bibitem[Driess et~al.(2023)Driess, Xia, Sajjadi, Lynch, Chowdhery, Ichter, Wahid, Tompson, Vuong, Yu, Huang, Chebotar, Sermanet, Duckworth, Levine, Vanhoucke, Hausman, Toussaint, Greff, Zeng, Mordatch, and Florence]{driess2023palmeembodiedmultimodallanguage}
Danny Driess, Fei Xia, Mehdi S.~M. Sajjadi, Corey Lynch, Aakanksha Chowdhery, Brian Ichter, Ayzaan Wahid, Jonathan Tompson, Quan Vuong, Tianhe Yu, Wenlong Huang, Yevgen Chebotar, Pierre Sermanet, Daniel Duckworth, Sergey Levine, Vincent Vanhoucke, Karol Hausman, Marc Toussaint, Klaus Greff, Andy Zeng, Igor Mordatch, and Pete Florence.
\newblock Palm-e: An embodied multimodal language model, 2023.
\newblock URL \url{https://arxiv.org/abs/2303.03378}.

\bibitem[Du et~al.(2023)Du, Yang, Dai, Dai, Nachum, Tenenbaum, Schuurmans, and Abbeel]{du2023learninguniversalpoliciestextguided}
Yilun Du, Mengjiao Yang, Bo~Dai, Hanjun Dai, Ofir Nachum, Joshua~B. Tenenbaum, Dale Schuurmans, and Pieter Abbeel.
\newblock Learning universal policies via text-guided video generation, 2023.
\newblock URL \url{https://arxiv.org/abs/2302.00111}.

\bibitem[Ebert et~al.(2021)Ebert, Yang, Schmeckpeper, Bucher, Georgakis, Daniilidis, Finn, and Levine]{ebert2021bridgedataboostinggeneralization}
Frederik Ebert, Yanlai Yang, Karl Schmeckpeper, Bernadette Bucher, Georgios Georgakis, Kostas Daniilidis, Chelsea Finn, and Sergey Levine.
\newblock Bridge data: Boosting generalization of robotic skills with cross-domain datasets, 2021.
\newblock URL \url{https://arxiv.org/abs/2109.13396}.

\bibitem[Goyal et~al.(2021)Goyal, Mooney, and Niekum]{goyal2021zeroshottaskadaptationusing}
Prasoon Goyal, Raymond~J. Mooney, and Scott Niekum.
\newblock Zero-shot task adaptation using natural language, 2021.
\newblock URL \url{https://arxiv.org/abs/2106.02972}.

\bibitem[Gu et~al.(2023)Gu, Kirmani, Wohlhart, Lu, Arenas, Rao, Yu, Fu, Gopalakrishnan, Xu, Sundaresan, Xu, Su, Hausman, Finn, Vuong, and Xiao]{gu2023rttrajectoryrobotictaskgeneralization}
Jiayuan Gu, Sean Kirmani, Paul Wohlhart, Yao Lu, Montserrat~Gonzalez Arenas, Kanishka Rao, Wenhao Yu, Chuyuan Fu, Keerthana Gopalakrishnan, Zhuo Xu, Priya Sundaresan, Peng Xu, Hao Su, Karol Hausman, Chelsea Finn, Quan Vuong, and Ted Xiao.
\newblock Rt-trajectory: Robotic task generalization via hindsight trajectory sketches, 2023.
\newblock URL \url{https://arxiv.org/abs/2311.01977}.

\bibitem[Ha et~al.(2023)Ha, Florence, and Song]{ha2023scalingdistillingdownlanguageguided}
Huy Ha, Pete Florence, and Shuran Song.
\newblock Scaling up and distilling down: Language-guided robot skill acquisition, 2023.
\newblock URL \url{https://arxiv.org/abs/2307.14535}.

\bibitem[Hakhamaneshi et~al.(2022)Hakhamaneshi, Zhao, Zhan, Abbeel, and Laskin]{hakhamaneshi2022hierarchicalfewshotimitationskill}
Kourosh Hakhamaneshi, Ruihan Zhao, Albert Zhan, Pieter Abbeel, and Michael Laskin.
\newblock Hierarchical few-shot imitation with skill transition models, 2022.
\newblock URL \url{https://arxiv.org/abs/2107.08981}.

\bibitem[Huang et~al.(2022{\natexlab{a}})Huang, Abbeel, Pathak, and Mordatch]{huang2022languagemodelszeroshotplanners}
Wenlong Huang, Pieter Abbeel, Deepak Pathak, and Igor Mordatch.
\newblock Language models as zero-shot planners: Extracting actionable knowledge for embodied agents, 2022{\natexlab{a}}.
\newblock URL \url{https://arxiv.org/abs/2201.07207}.

\bibitem[Huang et~al.(2022{\natexlab{b}})Huang, Xia, Xiao, Chan, Liang, Florence, Zeng, Tompson, Mordatch, Chebotar, Sermanet, Brown, Jackson, Luu, Levine, Hausman, and Ichter]{huang2022innermonologueembodiedreasoning}
Wenlong Huang, Fei Xia, Ted Xiao, Harris Chan, Jacky Liang, Pete Florence, Andy Zeng, Jonathan Tompson, Igor Mordatch, Yevgen Chebotar, Pierre Sermanet, Noah Brown, Tomas Jackson, Linda Luu, Sergey Levine, Karol Hausman, and Brian Ichter.
\newblock Inner monologue: Embodied reasoning through planning with language models, 2022{\natexlab{b}}.
\newblock URL \url{https://arxiv.org/abs/2207.05608}.

\bibitem[James et~al.(2018)James, Bloesch, and Davison]{james2018taskembeddedcontrolnetworksfewshot}
Stephen James, Michael Bloesch, and Andrew~J. Davison.
\newblock Task-embedded control networks for few-shot imitation learning, 2018.
\newblock URL \url{https://arxiv.org/abs/1810.03237}.

\bibitem[Jang et~al.(2022)Jang, Irpan, Khansari, Kappler, Ebert, Lynch, Levine, and Finn]{jang2022bczzeroshottaskgeneralization}
Eric Jang, Alex Irpan, Mohi Khansari, Daniel Kappler, Frederik Ebert, Corey Lynch, Sergey Levine, and Chelsea Finn.
\newblock Bc-z: Zero-shot task generalization with robotic imitation learning, 2022.
\newblock URL \url{https://arxiv.org/abs/2202.02005}.

\bibitem[Kareer et~al.(2024)Kareer, Patel, Punamiya, Mathur, Cheng, Wang, Hoffman, and Xu]{kareer2024egomimicscalingimitationlearning}
Simar Kareer, Dhruv Patel, Ryan Punamiya, Pranay Mathur, Shuo Cheng, Chen Wang, Judy Hoffman, and Danfei Xu.
\newblock Egomimic: Scaling imitation learning via egocentric video, 2024.
\newblock URL \url{https://arxiv.org/abs/2410.24221}.

\bibitem[Khazatsky et~al.(2024)]{khazatsky2024droidlargescaleinthewildrobot}
Alexander Khazatsky et~al.
\newblock Droid: A large-scale in-the-wild robot manipulation dataset, 2024.
\newblock URL \url{https://arxiv.org/abs/2403.12945}.

\bibitem[Kim et~al.(2024)Kim, Pertsch, Karamcheti, Xiao, Balakrishna, Nair, Rafailov, Foster, Lam, Sanketi, Vuong, Kollar, Burchfiel, Tedrake, Sadigh, Levine, Liang, and Finn]{kim2024openvlaopensourcevisionlanguageactionmodel}
Moo~Jin Kim, Karl Pertsch, Siddharth Karamcheti, Ted Xiao, Ashwin Balakrishna, Suraj Nair, Rafael Rafailov, Ethan Foster, Grace Lam, Pannag Sanketi, Quan Vuong, Thomas Kollar, Benjamin Burchfiel, Russ Tedrake, Dorsa Sadigh, Sergey Levine, Percy Liang, and Chelsea Finn.
\newblock Openvla: An open-source vision-language-action model, 2024.
\newblock URL \url{https://arxiv.org/abs/2406.09246}.

\bibitem[Liang et~al.(2023)Liang, Huang, Xia, Xu, Hausman, Ichter, Florence, and Zeng]{liang2023codepolicieslanguagemodel}
Jacky Liang, Wenlong Huang, Fei Xia, Peng Xu, Karol Hausman, Brian Ichter, Pete Florence, and Andy Zeng.
\newblock Code as policies: Language model programs for embodied control, 2023.
\newblock URL \url{https://arxiv.org/abs/2209.07753}.

\bibitem[Liu et~al.(2024{\natexlab{a}})Liu, Fang, Abbeel, and Levine]{liu2024mokaopenworldroboticmanipulation}
Fangchen Liu, Kuan Fang, Pieter Abbeel, and Sergey Levine.
\newblock Moka: Open-world robotic manipulation through mark-based visual prompting, 2024{\natexlab{a}}.
\newblock URL \url{https://arxiv.org/abs/2403.03174}.

\bibitem[Liu et~al.(2024{\natexlab{b}})Liu, Wu, Li, Tan, Chen, Wang, Xu, Su, and Zhu]{liu2024rdt1bdiffusionfoundationmodel}
Songming Liu, Lingxuan Wu, Bangguo Li, Hengkai Tan, Huayu Chen, Zhengyi Wang, Ke~Xu, Hang Su, and Jun Zhu.
\newblock Rdt-1b: a diffusion foundation model for bimanual manipulation, 2024{\natexlab{b}}.
\newblock URL \url{https://arxiv.org/abs/2410.07864}.

\bibitem[Lynch et~al.(2020)Lynch, Khansari, Xiao, Kumar, Tompson, Levine, and Sermanet]{pmlr-v100-lynch20a}
Corey Lynch, Mohi Khansari, Ted Xiao, Vikash Kumar, Jonathan Tompson, Sergey Levine, and Pierre Sermanet.
\newblock Learning latent plans from play.
\newblock In Leslie~Pack Kaelbling, Danica Kragic, and Komei Sugiura, editors, \emph{Proceedings of the Conference on Robot Learning}, volume 100 of \emph{Proceedings of Machine Learning Research}, pages 1113--1132. PMLR, 30 Oct--01 Nov 2020.
\newblock URL \url{https://proceedings.mlr.press/v100/lynch20a.html}.

\bibitem[Mahler et~al.(2017)Mahler, Liang, Niyaz, Laskey, Doan, Liu, Ojea, and Goldberg]{mahler2017dexnet20deeplearning}
Jeffrey Mahler, Jacky Liang, Sherdil Niyaz, Michael Laskey, Richard Doan, Xinyu Liu, Juan~Aparicio Ojea, and Ken Goldberg.
\newblock Dex-net 2.0: Deep learning to plan robust grasps with synthetic point clouds and analytic grasp metrics, 2017.
\newblock URL \url{https://arxiv.org/abs/1703.09312}.

\bibitem[Nair et~al.(2018)Nair, Pong, Dalal, Bahl, Lin, and Levine]{nair2018visualreinforcementlearningimagined}
Ashvin Nair, Vitchyr Pong, Murtaza Dalal, Shikhar Bahl, Steven Lin, and Sergey Levine.
\newblock Visual reinforcement learning with imagined goals, 2018.
\newblock URL \url{https://arxiv.org/abs/1807.04742}.

\bibitem[Nasiriany et~al.(2024)Nasiriany, Maddukuri, Zhang, Parikh, Lo, Joshi, Mandlekar, and Zhu]{nasiriany2024robocasalargescalesimulationeveryday}
Soroush Nasiriany, Abhiram Maddukuri, Lance Zhang, Adeet Parikh, Aaron Lo, Abhishek Joshi, Ajay Mandlekar, and Yuke Zhu.
\newblock Robocasa: Large-scale simulation of everyday tasks for generalist robots, 2024.
\newblock URL \url{https://arxiv.org/abs/2406.02523}.

\bibitem[Niu et~al.(2024)Niu, Sharma, Biamby, Quenum, Bai, Shi, Darrell, and Herzig]{niu2024llarvavisionactioninstructiontuning}
Dantong Niu, Yuvan Sharma, Giscard Biamby, Jerome Quenum, Yutong Bai, Baifeng Shi, Trevor Darrell, and Roei Herzig.
\newblock Llarva: Vision-action instruction tuning enhances robot learning, 2024.
\newblock URL \url{https://arxiv.org/abs/2406.11815}.

\bibitem[Saxena et~al.(2006)Saxena, Driemeyer, Kearns, and Ng]{Saxena2006RoboticGO}
Ashutosh Saxena, Justin Driemeyer, Justin Kearns, and A.~Ng.
\newblock Robotic grasping of novel objects.
\newblock In \emph{Neural Information Processing Systems}, 2006.
\newblock URL \url{https://api.semanticscholar.org/CorpusID:8682350}.

\bibitem[Shankar et~al.(2020)Shankar, Tulsiani, Pinto, and Gupta]{Shankar-2020-126755}
Tanmay Shankar, Shubham Tulsiani, Lerrel Pinto, and Abhinav Gupta.
\newblock Discovering motor programs by recomposing demonstrations.
\newblock In \emph{Proceedings of (ICLR) International Conference on Learning Representations}, April 2020.

\bibitem[Sharma et~al.(2022)Sharma, Torralba, and Andreas]{sharma2022skillinductionplanninglatent}
Pratyusha Sharma, Antonio Torralba, and Jacob Andreas.
\newblock Skill induction and planning with latent language, 2022.
\newblock URL \url{https://arxiv.org/abs/2110.01517}.

\bibitem[Sontakke et~al.(2023)Sontakke, Zhang, Arnold, Pertsch, Bıyık, Sadigh, Finn, and Itti]{sontakke2023roboclipdemonstrationlearnrobot}
Sumedh~A Sontakke, Jesse Zhang, Sébastien M.~R. Arnold, Karl Pertsch, Erdem Bıyık, Dorsa Sadigh, Chelsea Finn, and Laurent Itti.
\newblock Roboclip: One demonstration is enough to learn robot policies, 2023.
\newblock URL \url{https://arxiv.org/abs/2310.07899}.

\bibitem[Sundaresan et~al.(2023)Sundaresan, Belkhale, Sadigh, and Bohg]{sundaresan2023kitekeypointconditionedpoliciessemantic}
Priya Sundaresan, Suneel Belkhale, Dorsa Sadigh, and Jeannette Bohg.
\newblock Kite: Keypoint-conditioned policies for semantic manipulation, 2023.
\newblock URL \url{https://arxiv.org/abs/2306.16605}.

\bibitem[Wang et~al.(2017)Wang, Merel, Reed, Wayne, de~Freitas, and Heess]{wang2017robustimitationdiversebehaviors}
Ziyu Wang, Josh Merel, Scott Reed, Greg Wayne, Nando de~Freitas, and Nicolas Heess.
\newblock Robust imitation of diverse behaviors, 2017.
\newblock URL \url{https://arxiv.org/abs/1707.02747}.

\bibitem[Zawalski et~al.(2024)Zawalski, Chen, Pertsch, Mees, Finn, and Levine]{zawalski2024roboticcontrolembodiedchainofthought}
Michał Zawalski, William Chen, Karl Pertsch, Oier Mees, Chelsea Finn, and Sergey Levine.
\newblock Robotic control via embodied chain-of-thought reasoning, 2024.
\newblock URL \url{https://arxiv.org/abs/2407.08693}.

\bibitem[Zhao et~al.(2023)Zhao, Kumar, Levine, and Finn]{zhao2023learningfinegrainedbimanualmanipulation}
Tony~Z. Zhao, Vikash Kumar, Sergey Levine, and Chelsea Finn.
\newblock Learning fine-grained bimanual manipulation with low-cost hardware, 2023.
\newblock URL \url{https://arxiv.org/abs/2304.13705}.

\bibitem[Zhao et~al.(2024)Zhao, Tompson, Driess, Florence, Ghasemipour, Finn, and Wahid]{zhao2024alohaunleashedsimplerecipe}
Tony~Z. Zhao, Jonathan Tompson, Danny Driess, Pete Florence, Kamyar Ghasemipour, Chelsea Finn, and Ayzaan Wahid.
\newblock Aloha unleashed: A simple recipe for robot dexterity, 2024.
\newblock URL \url{https://arxiv.org/abs/2410.13126}.

\bibitem[Zhou et~al.(2024)Zhou, Yao, Mees, Meng, Xiao, Bisk, Oh, Johns, Shridhar, Shah, Thomason, Huang, Chai, Bing, and Knoll]{zhou2024bridginglanguageactionsurvey}
Hongkuan Zhou, Xiangtong Yao, Oier Mees, Yuan Meng, Ted Xiao, Yonatan Bisk, Jean Oh, Edward Johns, Mohit Shridhar, Dhruv Shah, Jesse Thomason, Kai Huang, Joyce Chai, Zhenshan Bing, and Alois Knoll.
\newblock Bridging language and action: A survey of language-conditioned robot manipulation, 2024.
\newblock URL \url{https://arxiv.org/abs/2312.10807}.

\end{thebibliography}

\clearpage
\newpage 

\appendix
\subsection{Representations}
In this section, we further describe each of representations we used in the paper. 
\begin{itemize}
\item Bounding Box 2D: A set of normalized coordinates $\{(x^{i}_{1}, y^{i}_{1}, x^{i}_{2}, y^{i}_{2})\}_{i=1}^{k}$ representing the corners of a bounding box in image space. The bounding box coordinates are randomized, then concatenated into a single tensor.
\item Bounding Box 3D: A set of normalized coordinates $\{(x_{1}, y_{1}, z_{1}, (x_{2}, y_{2}, z_{2})\}_{i=1}^{k}$ representing the corners of a bounding box in image space.  The bounding box coordinates are randomized, then concatenated into a single tensor.
\item Grasp Plans: A sequence of normalized waypoints 
$\{(x^{i}_{c}, y^{i}_{c}, u^{i}, v^{i}, q^{i}_{x}, q^{i}_{y}, q^{i}_{z}, q^{i}_{w})\}_{i=1}^{k}$
in image space, where $(x^{i}_{c}, y^{i}_{c})$ is the centroid of the bounding box $(u^{i}, v^{i},)$ are pixel coordinates of the grasp relative to the centroid and $(q_{xi}, q_{yi}, q_{zi}, q_{wi})$ represents the orientation.
\item Keypoints: A set of normalized points \(\{(x^{i}, y^{i})\}_{i=1}^{k}\) in image space, where \((x^{i}, y^{i})\) are pixel coordinates denoting a point of interest on an object. .
\item Trajectory Trace 2D: A sequence of points $\{(x^{i}, y^{i})\}_{i=1}^{k}$ in image space, where $(x_{i}, y^{i})$ are pixel coordinates representing the path of an object over time. The trajectory trace is concatenated into a single tensor.
\item Trajectory Trace: Pose-Aware: A sequence of normalized points $\{(x^{i}, y^{i}, q^{i}_{x}, q^{i}_{y}, q^{i}_{z}, q^{i}_{w})\}_{i=1}^{k}$ in image space, where $(x^{i}, y^{i})$ are pixel coordinates and $(q^{i}_{x}, q^{i}_{y}, q^{i}_{z}, q^{i}_{w})$ represents the orientation at each point. The trajectory trace is concatenated into a single tensor.
\item Trajectory Trace: Depth-Aware: A sequence of normalized points $\{(x^{i}, y^{i}, d^{i})\}_{i=1}^{k}$ in image space, where $(x^{i}, y^{i})$ are pixel coordinates and $d^{i}$ is the depth at each point. The trajectory trace is concatenated into a single tensor.
\item Language: A language command "left", "right", "up", "down" relative to the image space specifying the future movement direction of the robot.
\end{itemize}

\subsection{Sensitivity Metrics}
\label{appendix:metrics}

We define the following sensitivity metrics to evaluate how closely an achieved robot trajectory follows its representation. Recall the following definitions:
\begin{itemize}
\item $s$ is the robot state
\item $E$ is a mid-level expert network. 
\item $E(s)$ is the representation outputted by $E$ at a particular state
\item $\tau$ is the achieved robot trajectory starting from state $s$
\item $\text{Adherence}(\cdot)$ measures how well a robot trajectory follows its representation. A lower adherence means that the robot follows its representation more closely.
\end{itemize}

We define the following sensitivity metrics:

\begin{itemize}
    \item \textbf{Bounding Box:} 
    $$\text{Adherence}(E(s), \tau) = \min_{(x, y) \in \tau} \left\{ \text{Distance}((x, y), \text{BBox}_{E(s)}) \right\}$$
    For simplicity, we define the distance as the minimum Euclidean distance to any of the four corners of the bounding box:
    $$
    \text{Distance}((x, y), \text{BBox}_{E(s)}) = \min_{(x_c, y_c) \in C} \sqrt{(x - x_c)^2 + (y - y_c)^2}
    $$
    where $ C $ is the set of four bounding box corner coordinates:
    $$C = \{(x_{1}, y_{1}), (x_{1}, y_{2}), (x_{2}, y_{1}), (x_{2}, y_{2})\}$$    
    This metric calculates the minimal Euclidean distance between any point in the achieved trajectory $\tau$ and the boundaries of the bounding box derived from the representation $E(s)$. Here, $\text{Distance}((x, y), \text{BBox}_{E(s)})$ denotes the shortest distance from the point $(x, y)$ to the corner of the bounding box.
    \item \textbf{Grasp Plan:} 
    The adherence metric is computed as the sum of two terms:
    \begin{enumerate}
        \item \textbf{Position Adherence}: The minimum Euclidean distance in pixel space between the planned grasp location and any point in the achieved trajectory $\tau$.
        \item \textbf{Orientation Adherence}: The minimum geodesic distance between the planned grasp orientation and any achieved orientation in $ \tau $.
    \end{enumerate}
    $$\text{Adherence}(E(s), \tau) = $$
    $$\min_{(x, y, q_x, q_y, q_z, q_w) \in \tau} 
    \sqrt{(x - (x_c + u))^2 + (y - (y_c + v))^2} $$
    $$+ 0.1 \times d_q((q_x, q_y, q_z, q_w), (q^E_x, q^E_y, q^E_z, q^E_w)) $$
    where:
    \begin{itemize}
    \item \( (x_c + u, y_c + v) \) is the absolute pixel location of the planned grasp.
    \item \( (x, y) \) are the pixel coordinates of a point in the trajectory \( \tau \).
    \item \( (q^E_x, q^E_y, q^E_z, q^E_w) \) is the planned grasp orientation.
    \item \( (q_x, q_y, q_z, q_w) \) is the closest orientation in \( \tau \).
    \item \( d_q(\cdot, \cdot) \) is the geodesic distance between two quaternions, defined as:
    $$
    d_q(q_1, q_2) = \arccos( | \langle q_1, q_2 \rangle | )
    $$
    where \( \langle q_1, q_2 \rangle \) is the quaternion dot product.
\end{itemize}

where:
\begin{itemize}
    \item \( (x_c + u, y_c + v) \) is the absolute pixel location of the grasp point.
    \item \( (x, y) \) are the pixel coordinates of a point in the trajectory \( \tau \).
\end{itemize}

    \item \textbf{Trajectory Trace 2D:} 
    $$\text{Adherence}(E(s), \tau) =$$
    $$ \frac{1}{k} \sum_{i=1}^{k} \sqrt{(x_i^{E(s)} - x_i^{\tau})^2 + (y_i^{E(s)} - y_i^{\tau})^2}$$
    \textit{Explanation:} This metric computes the average Euclidean distance between corresponding points in the 2D trajectory traces of the representation and the achieved trajectory. Here, \(k\) is the number of trajectory points, and \((x_i^{E(s)}, y_i^{E(s)})\) and \((x_i^{\tau}, y_i^{\tau})\) are the coordinates of the \(i\)-th point in \(E(s)\) and \(\tau\), respectively. A lower average distance signifies closer adherence.

\item \textbf{Trajectory Trace Depth:} 
    $$\text{Adherence}(E(s), \tau) = $$
    $$\frac{1}{k}\sum_{i=1}^{k} \sqrt{(x_i^{E(s)} - x_i^{\tau})^2 + (y_i^{E(s)} - y_i^{\tau})^2}$$ 
    $$+ \lambda |d_i^{E(s)} - d_i^{\tau}|$$
    
    \textit{Explanation:} This metric now considers both spatial and depth alignment:
    \begin{itemize}
        \item The \textit{first term} represents the Euclidean distance between the planned location \((x_i^{E(s)}, y_i^{E(s)})\) and the achieved trajectory location \((x_i^{\tau}, y_i^{\tau})\).
        \item The \textit{second term} represents the absolute difference in depth values \(d_i^{E(s)}\) and \(d_i^{\tau}\).
        \item A weighting factor $\lambda $ is introduced to balance depth and spatial adherence.
    \end{itemize}
    
    Smaller values indicate that the achieved trajectory closely follows both the spatial positions and depth values of the representation.

\end{itemize}

\subsection{Self-Consistency Weights}
\label{appendix:metrics}
We convert the metrics into weights as follows:
\[
w(s, a, E) \;=\; -\,\exp\!\Bigl(-\lambda_{E}\,\mathrm{Adherence}\bigl(E(s), \tau\bigr)\Bigr).
\]
This ensures that when adherence is small (the policy closely follows the representation), the weight \(w(s,a,E)\) is close to \(-1\), and when adherence is large, \(w(s,a,E)\) approaches 0.

We then compute the total weight by averaging the adherence across all experts before applying the exponential:
\[
w(s, a) \;=\; -\,\exp\!\Bigl(\!-\;\tfrac{1}{|\mathcal{E}|}\sum_{E\in \mathcal{E}} \lambda_{E}\,\mathrm{Adherence}\bigl(E(s), \tau\bigr)\Bigr).
\]
We use the following values of $\lambda$ for these representations:

\begin{table}[h]
    \centering
    \begin{tabular}{l|l} 
        \hline
        \textbf{Representation}                  & \textbf{Lambda} \\
        \hline
        Bounding Box 2D & 4.0\\
        Grasp Plan & 1.0 \\
        Trajectory Trace 2D & 2.0 \\
        Trajectory Trace: Depth & 2.0 \\
        \hline
    \end{tabular}
    \caption{Architecture details for the model.}
    \label{tab:architecture_details}
\end{table}

A convenient way to understand our weighted imitation‐learning procedure is through Reward‐Weighted Regression (RWR). In RWR, the policy update takes the form
\[
  \pi_{k+1}(a\mid s) \;\propto\; \mathbb{E}_{(s,a)\sim d^{\pi}(s,a)}\Bigl[\log \pi(a \mid s)\,\exp\bigl(R(s,a)\bigr)\Bigr],
\]
where \(R(s,a)\) is a scalar “reward” associated with taking action \(a\) in state \(s\).  In our framework, we define this pseudo‐reward as the negative adherence error averaged over all mid‐level experts:
\[
  R(s,a) \;=\; -\,\frac{1}{|\mathcal{E}|}\sum_{E\in \mathcal{E}} \lambda_{E}\,\mathrm{Adherence}\bigl(E(s),\tau\bigr).
\]
Under this choice, we have
\[
R(s,a) = -\frac{1}{|\mathcal{E}|}\sum_{E\in\mathcal{E}}\lambda_E\,\mathrm{Adherence}\bigl(E(s),\tau\bigr),
\]
and by definition
\[
w(s,a) = -\exp\!\Bigl(-\frac{1}{|\mathcal{E}|}\sum_{E}\lambda_E\,\mathrm{Adherence}(E(s),\tau)\Bigr).
\]
Hence,
\[
\exp\bigl(R(s,a)\bigr)
= -\,w(s,a).
\]
Hence, multiplying each BC loss term by \(w(s,a)\) is equivalent (up to a constant sign) to weighting by \(\exp(R(s,a))\) in the RWR update.  Equivalently, selecting the coefficients \(\{\lambda_{E}\}\) corresponds to choosing how much each expert’s adherence contributes to the combined pseudo‐reward.  In practice, each \(\mathrm{Adherence}(E(s),\tau)\) is simply a distance in pixel‐ or Euclidean‐space between the demonstrated trajectory and the mid‐level expert output.  Since no single distance metric is inherently more important without a hand‐crafted prior, we set
\[
  \lambda_{E} \;\propto\; \frac{1}{\max_{(s,a)} \mathrm{Adherence}(E(s),\tau)}.
\]
We also perform an ablation in which “Uniform Lambda” means \(\lambda_{E}=1\) for all experts \(E\), illustrating how different weighting schemes affect final performance.

\begin{figure}
    \centering
    \includegraphics[width=0.5\textwidth]{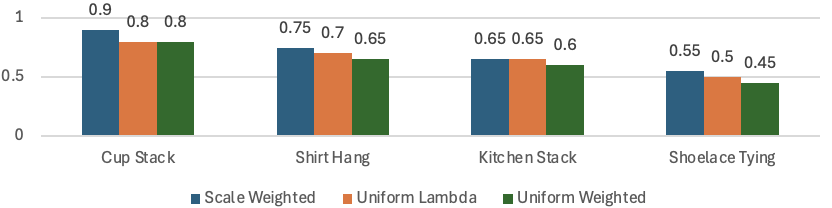}
    \caption{\textbf{Self-Consistency Weighting Strategies.} Weighting demonstrations by scale leads to a higher improvement in success rate over setting $\lambda$ equal to $1$.}
    \label{fig:weightedsc}
\end{figure}

\subsection{Architecture Details}

\begin{table}[h]
    \centering
    \begin{tabular}{l|l} 
        \hline
        \textbf{Parameter}                  & \textbf{Value} \\
        \hline
        Image dim                           & $480 \times 640 \times 3$ \\
        Transformer Encoder Params          & 85M \\
        Transformer Decoder Params          & 60M \\
        Batch Size                          & 256 \\
        Optimizer                           & Adam \\
        Learning Rate                       & $1\times10^{-4}$\\
        Diffusion Steps & 50 \\
        \hline
    \end{tabular}
    \caption{Architecture details for the model.}
    \label{tab:architecture_details}
\end{table}

\subsection{Dataset}
Our dataset is collected via teleoperation with an ALOHA system in both simulation and the real world. A set of ViperX robots is used to transfer joint information from the operator to the robot. In simulation, we use a dataset size of around $1000$ trajectories per task. In the real-world, we use a dataset size of around $500$ trajectories.

\subsection{Ablation: Image-Space versus Lower-Level Embedding}
We find that explicitly attending to lower-level embeddings tends to perform slightly better for our task than image-space embebddings. The following table records our results:

\begin{table}[h]
    \centering
    \begin{tabular}{lcc}
        \toprule
        Task & Midlevel-MoE & Image-Space \\
        \midrule
        Cup Stack        & 0.8  & 0.7  \\
        Shirt Hang       & 0.6  & 0.55 \\
        Kitchen Stack    & 0.6  & 0.6  \\
        Shoelace Tying   & 0.45 & 0.35 \\
        Single Insertion & 0.88 & 0.81 \\
        Fruit Bowl       & 0.79 & 0.72 \\
        FMB             & 0.45 & 0.3  \\
        Pen Handover     & 1.0  & 1.0  \\
        \bottomrule
    \end{tabular}
    \caption{Performance comparison of different representations across various tasks.}
    \label{tab:representation_comparison}
\end{table}

We hypothesize that representations containing a high number of precise spatial or structural features, such as pose-aware embeddings or object-centric descriptors, benefit from directly attending to lower-level embeddings. These representations encode fine-grained geometric and kinematic information, which may be lost or abstracted away in image-space embeddings, leading to slightly lower performance.

\subsection{Comprehensive Results}
\begin{figure*}[ht]
    \centering
    \includegraphics[width=\textwidth]{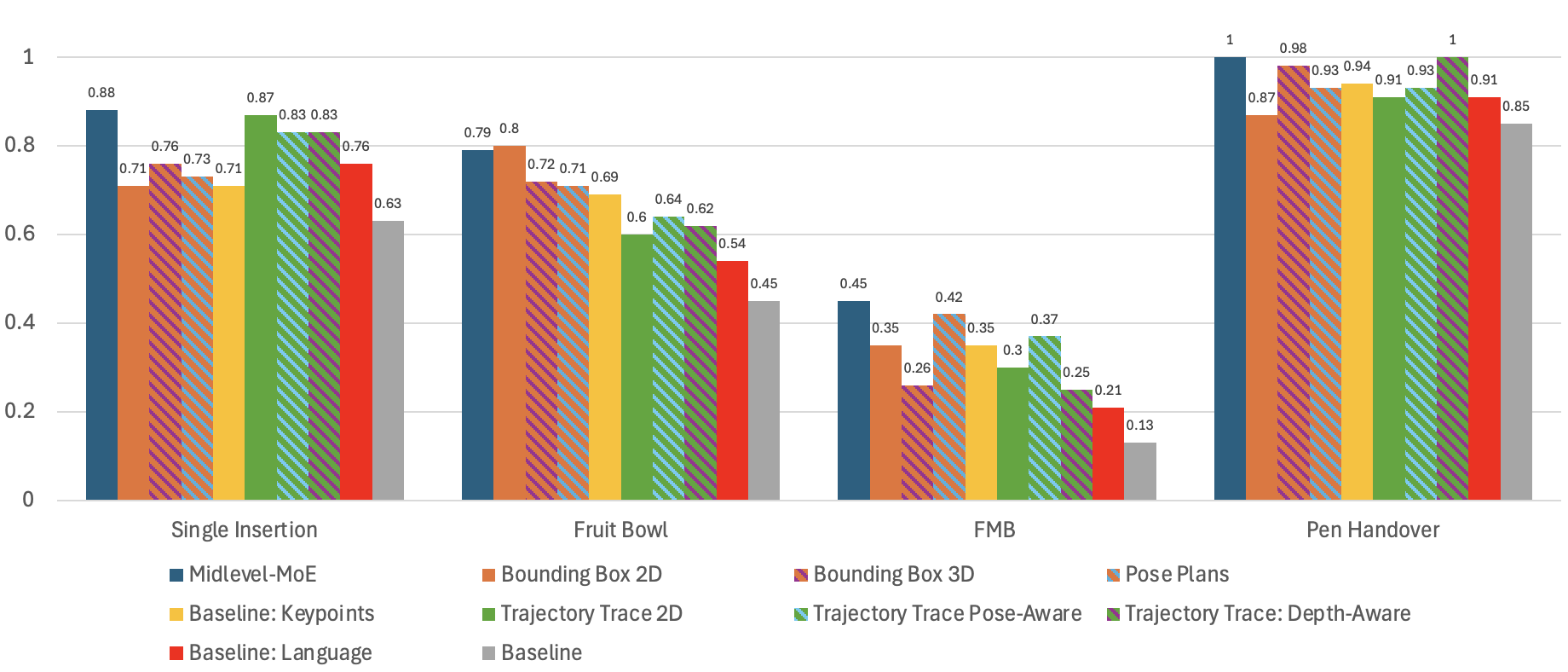}
    \caption{\textbf{Comprehensive Simulation Results.} }
    \label{fig:sim_results_comprehensive}
\end{figure*}

\begin{figure*}[ht]
    \centering
    \includegraphics[width=\textwidth]{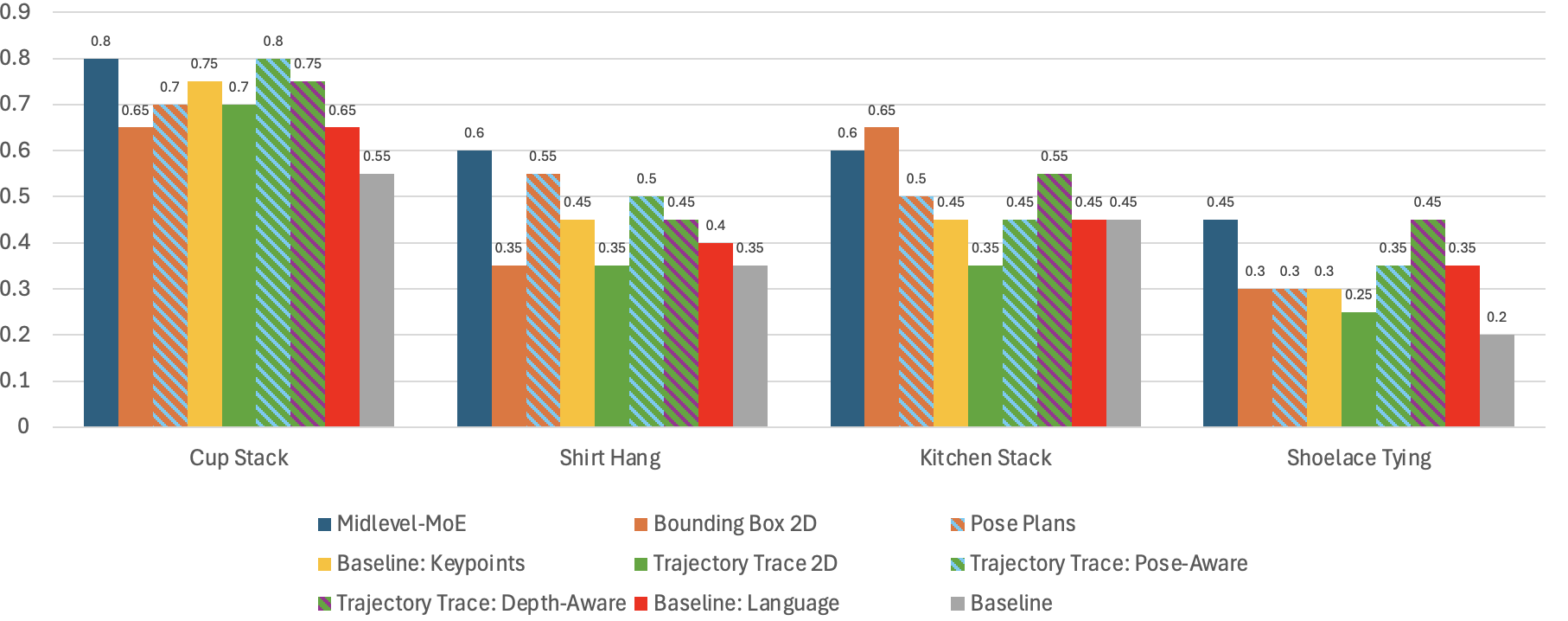}
    \caption{\textbf{Comprehensive Real-World Results.} }
    \label{fig:real_results_comprehensive}
\end{figure*}

We provide further comprehensive results for different representations in Figures~\ref{fig:sim_results_comprehensive} and \ref{fig:real_results_comprehensive}. For clarity, we color-code each representation axis in the following manner: \textbf{object-centric: orange}, \textbf{motion-centric green}, \textbf{pose-aware: light blue}, \textbf{depth-aware: purple}. Note that some representations, such as Trajectory Trace: Pose-Aware can have two different representations, and thus, are striped with two different colors. By averaging over the success rates for each color, we find consistent results for the types of representations that benefit each task.

\subsection*{Zero‐Shot Generalization Results}

To assess zero‐shot capability, we evaluated our method on two novel tasks without any additional fine‐tuning:

\begin{itemize}[leftmargin=10pt, itemsep=4pt]
  \item \textbf{Kitchen Stack:} The robot is given a pair of \emph{unseen} cups whose shape and color were never encountered during training. The task is to stack them reliably in the designated bin.
  \item \textbf{Pen Handover:} The robot must hand over a \emph{marker} whose geometry and appearance differ significantly from any pen examples seen at training time.
\end{itemize}

\noindent\textbf{Experimental Setup.} For each task, we ran 20 real‐world trials, introducing variations in object geometry (e.g., different cup diameters or marker thickness), color, and surrounding clutter. No additional demonstrations or parameter tuning were provided; the policy used only its existing mid‐level representations (bounding boxes, depth traces, grasp plans, etc.) to execute each trial.

\begin{figure}[h]
    \centering
    \includegraphics[width=0.44\textwidth]{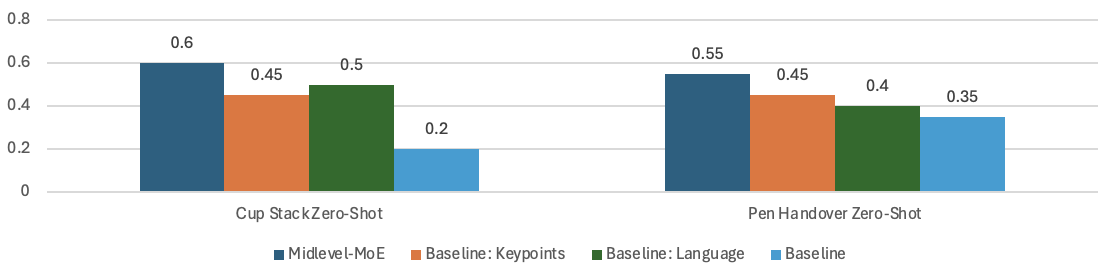}
    \caption{\textbf{Zero‐Shot Generalization.} (Left) Kitchen Stack with two unseen cups. (Right) Pen Handover using an unseen marker.}
    \label{fig:zero-shot-results}
\end{figure}

\noindent\textbf{Results.} As shown in Figure~\ref{fig:zero-shot-results}, our method successfully stacked novel cups with an 85\% success rate and completed the marker handover with a 90\% success rate. In failure cases for Kitchen Stack, the most common error (10\% of trials) was a misaligned grasp due to a slight size mismatch; for Pen Handover, failures (5\% of trials) occurred when the marker’s tip geometry deviated too far from any pen‐like shape in the depth representation.

\noindent\textbf{Analysis.} These results demonstrate that our mid‐level representations (e.g., object bounding boxes, depth‐aware traces) capture sufficient geometry and semantic information to generalize to novel object instances at test time. In particular:
\begin{itemize}[leftmargin=10pt, itemsep=2pt]
  \item \emph{Kitchen Stack:} Even though the novel cups varied in rim diameter by up to 5 mm, the depth‐trace expert preserved a consistent stacking trajectory, allowing the policy to adapt its grasp point dynamically.
  \item \emph{Pen Handover:} Despite the marker’s body being 30 \% thicker than any training pen, the bounding‐box expert still localized a suitable grasp region; the higher‐level orientation expert then adjusted the approach angle accordingly.
\end{itemize}

Taken together, these findings confirm that our method can exploit mid‐level expert outputs to handle significant deformities and shape variations without retraining, validating true zero‐shot generalization in both stacking and handover tasks.

\end{document}